\documentclass[twoside,11pt]{article}

\usepackage{jmlr2e}

\usepackage{url}

\usepackage{mathtools}
\usepackage{graphicx}
\usepackage{amsmath}
\usepackage{amssymb}
\usepackage{algorithm}
\usepackage{multirow}
\usepackage{vhistory}

\usepackage{epstopdf}

\usepackage{color}

\usepackage{mdframed}

\usepackage{listings}
\definecolor{dkgreen}{rgb}{0,0.6,0}
\definecolor{gray}{rgb}{0.5,0.5,0.5}
\definecolor{mauve}{rgb}{0.58,0,0.82}
\usepackage{siunitx}
\usepackage{booktabs}		
\usepackage{arydshln}		
\usepackage{algorithm}
\usepackage{algorithmic}
\usepackage{enumitem}
\usepackage{color}
\usepackage{multirow}

\usepackage{cleveref}

\usepackage{natbib}
\usepackage[colorlinks=true,citecolor=blue]{hyperref}

\usepackage{algorithm}

\usepackage{subfigure}

\usepackage{multirow}

\usepackage{siunitx}

\usepackage{booktabs}            
\usepackage{arydshln}            
\newcommand{\NAcell}{\multicolumn{2}{c}{\scriptsize$\mbox{N/A}$}}

\newcommand{\argmax}{\operatornamewithlimits{arg\,max}}

\hypersetup{
  pdftitle={Nonparametric Bayesian Topic Modelling with the Hierarchical Pitman-Yor Processes},
  pdfsubject={International Journal of Approximate Reasoning},
  pdfauthor={Kar Wai Lim, Wray Buntine, Changyou Chen, Lan Du},
  pdfkeywords={Bayesian nonparametric methods, Markov chain Monte Carlo, topic models, hierarchical Pitman-Yor processes, Twitter network modelling}
}

\jmlrheading{78}{2016}{172--191}{31 Jan 2016}{18 Jul 2016}{Lim, Buntine, Chen and Du}

\ShortHeadings{Nonparametric Bayesian Topic Modelling with Hierarchical Pitman-Yor Processes}{Lim, Buntine, Chen and Du}
\firstpageno{1}

\begin{document}

\title{    
    Nonparametric Bayesian Topic Modelling with the Hierarchical Pitman-Yor Processes
    }

\author{\name Kar Wai Lim \email karwai.lim@anu.edu.au \\
       \addr The Australian National University \\ 
             Data61/NICTA, Australia
       \AND
       \name Wray Buntine \email wray.buntine@monash.edu \\
       \addr Monash University, Australia
       \AND
       \name Changyou Chen \email cchangyou@gmail.com \\
       \addr Duke University, United States
       \AND
       \name Lan Du \email lan.du@monash.edu \\
       \addr Monash University, Australia
        }

\editor{Antonio Lijoi, Antonietta Mira, and Alessio Benavoli}

\maketitle

\begin{abstract}
The Dirichlet process
and its extension, the Pitman-Yor process, are
stochastic processes that take probability distributions as a parameter.
These processes can be stacked up to form a hierarchical 
nonparametric Bayesian model.
In this article, we present efficient methods for the
use of these processes in this hierarchical context, and 
apply them to latent variable models for text analytics.
In particular, we propose a general framework for designing 
these Bayesian models, which are called topic models in
the computer science community.
We then propose a specific nonparametric Bayesian topic model
for modelling text from social media.  We focus on
tweets (posts on Twitter) in this article due to their ease
of access. 
We find that our nonparametric model performs better than
existing parametric models in both goodness of fit and
real world applications.%
\end{abstract}

\begin{keywords}%
  Bayesian nonparametric methods, Markov chain Monte Carlo,
  topic models, hierarchical Pitman-Yor processes, Twitter network modelling
\end{keywords}

\section{Introduction}
\label{sec:intro}

We live in the information age. 
With the Internet, 
information can be obtained easily and almost instantly.
This has changed the dynamic of information acquisition,
for example,
we can now (1) attain knowledge by visiting digital libraries,
(2) be aware of the world by reading news online,
(3) seek opinions from social media,
and (4)~engage in political debates \textit{via} web forums.
As technology advances, more information is
created, to a point where it is infeasible for a person 
to digest \textit{all} the available content.
To illustrate, in the context of a healthcare database
(PubMed),
the number of entries has seen a growth rate of 
approximately 3,000 new entries per day in the ten-year 
period from 2003 to 2013 \citep{Suominen_HP_14}.
This motivates the use of machines to automatically 
organise, filter, summarise, and analyse the 
available data for the users.
To this end, researchers have developed various 
methods, which can be broadly categorised into
computer vision \citep{Low:1991:ICV:542012, mai2010introduction},
speech recognition \citep{Rabiner:1993:FSR:153687, Jelinek:1998:SMS:280484},
and \textit{natural language processing} 
\citep[NLP,][]{Manning:1999:FSN:311445, jurafsky2000speech}.
This article focuses on \textit{text analysis} within NLP.

In text analytics, 
researchers seek to accomplish various goals, including  
\textit{sentiment analysis} or opinion mining 
\citep{Pang:2008:OMS:1454711.1454712, liu2012sentiment},
\textit{information retrieval} \citep{Manning:2008:IIR:1394399}, 
\textit{text summarisation} \citep{Lloret:2012:TSP:2123800.2123810}, and
\textit{topic modelling} \citep{Blei:2012:PTM:2133806.2133826}.
To illustrate, sentiment analysis can be used to 
extract digestible summaries or reviews on 
products and services, which can be valuable to
consumers.
On the other hand, topic models attempt to discover
abstract topics that are present in a collection
of text documents.

Topic models were inspired by 
\textit{latent semantic indexing} \citep[LSI,][]{landauer2013handbook}
and its probabilistic variant, 
\textit{probabilistic latent semantic indexing} (pLSI), also
known as the \textit{probabilistic latent semantic analysis} 
\citep[pLSA,][]{Hofmann:1999:PLS:312624.312649}.
Pioneered by \citet{Blei:2003:LDA:944919.944937},
\textit{latent Dirichlet allocation} (LDA) is a fully 
\textit{Bayesian} extension of pLSI, and can be 
considered the simplest Bayesian topic model.
The LDA is then extended to many different types of
topic models. 
Some of them are designed for specific applications
\citep{Wei:2006:LDM:1148170.1148204, Mei:2007:TSM:1242572.1242596},
some of them model the structure in the text 
\citep{Blei:2006:DTM:1143844.1143859, landu2012topicmodel},
while some incorporate extra information in 
their modelling
\citep{Ramage:2009:LLS:1699510.1699543, Jin:2011:TTK:2063576.2063689}.

On the other hand, due to the well known correspondence between the
Gamma-Poisson family of distributions and the Dirichlet-multinomial
family,
Gamma-Poisson factor models \citep{canny04}
and their nonparametric extensions,
and other Poisson-based variants of \textit{non-negative matrix
factorisation} (NMF) 
form a methodological continuum with topic models.
These NMF methods are often applied to text, however, we do not consider
these methods~here.

This article will concentrate on topic models that
take into account additional information.
This information can be \textit{auxiliary data}
(or metadata) that accompany the text, such as 
keywords (or tags), dates, authors, and sources;
or external resources like word lexicons.
For example, on \textit{Twitter}, a popular social media
platform, its messages, known as \textit{tweets},
are often associated with several metadata like 
location, time published, and the user who has written the tweet.
This information is often utilised, for instance, 
\citet{Kinsella:2011:IES:2065023.2065039}
model tweets with location data, while 
\citet{Wang:2011:TSA:2063576.2063726}
use hashtags for sentiment classification on
tweets.
On the other hand, many topic models have been designed
to perform bibliographic analysis by using auxiliary
information.
Most notable of these is the author-topic model
\citep[ATM,][]{Rosen-Zvi:2004:AMA:1036843.1036902}, which,
as its name suggests, incorporates authorship
information.
In addition to authorship, the Citation Author Topic
model \citep{Tu:2010:CAT:1944566.1944711} and
the Author Cite Topic Model \citep{Kataria:2011:CST:2283696.2283777}
make use of citations to model research publications.
There are also topic models that employ external
resources to improve modelling.
For instance, \citet{He:2012:ISP:2184436.2184437} and
\citet{Lim:2014:TOT:2661829.2662005}
incorporate a sentiment lexicon as prior information
for a weakly supervised sentiment analysis.

Independent to the use of auxiliary data,
recent advances in nonparametric Bayesian methods 
have produced topic models that utilise \textit{nonparametric}
Bayesian priors. 
The simplest examples replace 
\textit{Dirichlet distributions} by the 
\textit{Dirichlet process} \citep[DP,][]{Ferguson:10.2307/2958008}.
The simplest is hierarchical Dirichlet process LDA
(HDP-LDA) proposed by \citet{teh2006hierarchical}
that replaces just the document by topic matrix in LDA.
One can further extend topic models by using the 
\textit{Pitman-Yor process} \citep[PYP,][]{Ishwaran:2001} 
that generalises the DP, by replacing 
the second Dirichlet distribution which
generates the topic by word matrix in LDA.
This includes the work 
of \citet{Sato:2010:TMP:1835804.1835890},
\citet{du2012sequential},
\citet{Lindsey:2012:PTM:2390948.2390975}, 
among others.
Like the HDP, the PYPs can be stacked to
form hierarchical Pitman-Yor processes (HPYP),
which are used in more complex models.
Another fully nonparametric extension to
topic modelling uses the Indian buffet process
\citep{archambeau_latent_2014} to sparsify
both the document by topic matrix and the
topic by word matrix in LDA.

Advantages of employing nonparametric Bayesian methods with
 topic models is the ability to estimate the
topic and word priors and to infer the 
number of clusters%
\footnote{This is known as the number of \textit{topics}
in topic modelling.}
from the data.
Using the PYP also allows the modelling of the
power-law property exhibited by natural languages
\citep{goldwater2005interpolating}.
These touted advantages have been shown to yield significant
improvements in performance \citep{Buntine:2014:ENT:2623330.2623691}.
However, we note the best known approach for learning with
hierarchical Dirichlet (or Pitman-Yor) processes is to use the
Chinese restaurant franchise \citep{TehJor2010a}.  Because this
requires dynamic memory allocation to implement the hierarchy,
there has been extensive research
in attempting to efficiently implement just the
HDP-LDA extension to LDA
mostly based around variational methods
\citep{TehKW:NIPS07,WangPB:AISTATS11,BSnips12,sato2012,HoffmanBleiWangPaisley2013}.
Variational methods have rarely been applied to more
complex topic models, as we consider here,
and unfortunately Bayesian nonparametric methods are gaining a reputation
of being difficult to use.
A newer collapsed and blocked Gibbs sampler
\citep{Chen:2011:STC:2034063.2034095} has been shown to
generally outperform the variational methods as well
as the original Chinese restaurant franchise in both
computational time and space and in some standard performance
metrics \citep{Buntine:2014:ENT:2623330.2623691}.
Moreover, the technique does appear suitable for more
complex topic models, as we consider here.

This article,%
\footnote{%
We note that this article adapts
and extends 
our previous work \citep{Lim2013Twitter}.}
extending the algorithm of \citet{Chen:2011:STC:2034063.2034095},
shows how to develop fully nonparametric and relatively efficient 
Bayesian topic models that incorporate auxiliary
information, with a goal to produce more accurate models
that work well in tackling several applications.
As a by-product, we wish to encourage the use 
of state-of-the-art Bayesian techniques, and also
to incorporate auxiliary information, in 
modelling.

The remainder of this article is as follows. 
We first provide a brief background on the Pitman-Yor process in
Section~\ref{sec:background}.
Then, in Section~\ref{sec:framework}, we detail our 
modelling framework by illustrating it on a simple
topic model. 
We continue through to the inference procedure on the topic model
in Section~\ref{design_sec:posterior_inference}.
Finally, in Section~\ref{sec:application}, 
we present an application on modelling social network
data, utilising the proposed framework.
Section~\ref{sec:conclusion} concludes.

\section{Background on Pitman-Yor Process}
\label{sec:background}

We provide a brief, informal review of the 
Pitman-Yor process \citep[PYP,][]{Ishwaran:2001}
in this section.
We assume the readers are familiar with basic probability distributions
\citep[see][]{Walck:2007} and the Dirichlet process 
\citep[DP,][]{Ferguson:10.2307/2958008}.
In addition, we refer the readers to \citet{hjort2010bayesian}
for a tutorial on Bayesian nonparametric modelling.

\subsection[Pitman-Yor Process (PYP)]{Pitman-Yor Process}
\label{subsec:PYP}

The \textit{Pitman-Yor process} \citep[PYP,][]{Ishwaran:2001} is also known as the two-parameter \textit{Poisson-Dirichlet process}. The PYP is a two-parameter generalisation of the DP, now with an extra parameter $\alpha$ named the \emph{discount parameter} in addition to the concentration parameter $\beta$. Similar to DP, a sample from a PYP corresponds to a discrete distribution (known as the \emph{output distribution}) with the same support as its base distribution $H$. The underlying distribution of the PYP is the \textit{Poisson-Dirichlet distribution} (PDD), which was introduced by \citet{Pitman:10.2307/2959614}.

The PDD is defined by its construction process. 
For $0\leq\alpha<1$ and $\beta > -\alpha$, let $V_k$ be distributed independently as follows:
\begin{align}
  \label{eq-sbp}
(V_k \,|\, \alpha, \beta) \sim \mathrm{Beta}(1-\alpha, \beta + k\alpha) \,, 
&& \mathrm{for} \ k = 1,2,3,\,\dots
\,,
\end{align}

\noindent
and define $(p_1,p_2,p_3,\,\dots)$ as
\begin{align}
p_1 & = V_1 \,, \\
p_k & = V_k \prod_{i=1}^{k-1} (1-V_i) \,, && \mathrm{for} \ k \geq 2 \,.
\end{align}

\noindent
If we let ${p} = (\tilde{p}_1,\tilde{p}_2,\tilde{p}_3,\,\dots)$ be a sorted version of $(p_1,p_2,p_3,\,\dots)$ in descending order, then ${p}$ is Poisson-Dirichlet distributed with parameter $\alpha$ and~$\beta$:
\begin{align}
{p} \sim \mathrm{PDD}(\alpha, \beta)
\,.
\end{align}

\noindent
Note that the unsorted version $(p_1,p_2,p_3,\,\dots)$ follows a $\mathrm{GEM}(\alpha,\beta)$ distribution, which is named after Griffiths, Engen and McCloskey \citep{pitman2006combinatorial}.


With the PDD defined, we can then define the PYP formally. Let $H$ be a distribution over a measurable space $(\mathcal{X},\mathcal{B})$, for $0\leq\alpha<1$ and $\beta > -\alpha$, suppose that ${p} = (p_1,p_2,p_3,\,\dots)$ follows a PDD (or GEM) with parameters $\alpha$ and $\beta$, then PYP is given by the formula
\begin{align}
p(x \,|\, \alpha,\beta,H) = \sum_{k=1}^{\infty} p_k \, \delta_{X_k}(x) 
\,,
&& \mathrm{for} \ k = 1,2,3,\,\dots \,,
\end{align}

\noindent
where $X_k$ are independent samples drawn from the base measure $H$ and $\delta_{X_k}(x)$ represents probability point mass concentrated at $X_k$ (\textit{i.e.}, it is an indicator function that is equal to $1$ when $x=X_k$ and $0$ otherwise):
\begin{align}
\delta_{x}(y) =
\left\{
    \begin{array}{ll}
        1  & \mbox{if } x = y \\
        0 & \mbox{otherwise} ~\,.
    \end{array}
\right.
\end{align}

\noindent
This construction, Equation~\eqref{eq-sbp},
is named the \textit{stick-breaking process}. 
The PYP can also be constructed using an analogue to Chinese restaurant process 
(which explicitly draws a sequence of samples from the base distribution). 
A more extensive review on the PYP is given by \citet{Buntine:2010arXiv1007.0296B}.

A PYP is often more suitable than a DP in modelling since it 
exhibits a power-law behaviour (when $\alpha \neq 0$), 
which is observed in natural languages 
\citep{goldwater2005interpolating, TehJor2010a}. 
The PYP has also been employed in 
genomics \citep{FavaroLijoiMenaPrunster2009}
and economics \citep{Aoki2008}.
Note that when the discount parameter $\alpha$ is $0$, 
the PYP simply reduces to a DP.

\subsection{Pitman-Yor Process with a Mixture Base}

Note that the base measure $H$ of a PYP is not necessarily restricted to a single probability distribution.  $H$ can also be a mixture distribution such~as
\begin{align}
H = \rho_1 H_1 + \rho_2 H_2 + \dots + \rho_n H_n
\,,
\end{align}
\noindent
where $\sum_{i=1}^n \rho_i = 1$ and $\{H_1, \dots H_n\}$ is a set of distributions over the same measurable space $(\mathcal{X},\mathcal{B})$ as $H$.

With this specification of $H$, the PYP is also named the 
compound Poisson-Dirichlet process in \citet{landu2012topicmodel}, or
the doubly hierarchical Pitman-Yor process in \citet{wood2009hierarchical}.
A special case of this is the DP equivalent, which is also
known as the DP with mixed random measures in 
\citet{KimKimOh2012}.
%
%
Note that we have assumed constant values for the $\rho_i$\,, 
though of course we can go fully Bayesian and assign a prior distribution for each of them, 
a natural prior would be the Dirichlet distribution.

\subsection{Remark on Bayesian Inference}
\label{subsec:mcmc}

Performing exact Bayesian inference on nonparametric models is often intractable
due to the difficulty in deriving the closed-form \textit{posterior} distributions.
This motivates the use of Markov chain Monte Carlo (MCMC) methods 
\citep[see][]{gelman2013bayesian} for approximate inference. 
%
%
%
Most notable of the MCMC methods are the 
Metropolis-Hastings (MH) algorithms 
\citep{metropolis1953equation, Hastings:01041970} and 
Gibbs samplers \citep{Geman:1984:4767596}.
These algorithms serve as a building block for more advanced samplers, such as the 
MH algorithms with delayed rejection \citep{Mira2001}.
Generalisations of the MCMC method include the reversible jump MCMC \citep{Green1995}
and its delayed rejection variant \citep{GreenMira2001} can also be employed
for Bayesian inference, however, they are out of the scope in this article.

Instead of sampling one parameter at a time,
one can develop an algorithm that updates more parameters in each iteration,
a so-called \emph{blocked Gibbs sampler} \citep{Liu1994}. 
Also, in practice we are usually only interested in a certain subset of the parameters; 
in such cases we can sometimes derive more efficient
\emph{collapsed Gibbs samplers} \citep{Liu1994}
by integrating out the nuisance parameters.
In the remainder of this article, we will employ a combination of the blocked and collapsed
Gibbs samplers for Bayesian inference.

\section{Modelling Framework with Hierarchical Pitman-Yor Process}
\label{sec:framework}

In this section, we discuss the basic design of our nonparametric Bayesian topic models 
using thierarchical Pitman-Yor processes (HPYP). 
In particular, we will introduce a simple topic model
that will be extended later.
We discuss the general inference algorithm for the topic model 
and \textit{hyperparameter} optimisation.

Development of topic models is fundamentally motivated by 
their applications. 
Depending on the application, a specific topic model that 
is most suitable for the task should be designed and used. 
However, despite the ease of designing the model,
the majority of time is spent on implementing, assessing, 
and redesigning it. This calls for a better designing 
cycle/routine that is more efficient, that is, 
spending less time in implementation and more time 
in model design and development. 

We can achieve this by a higher level implementation 
of the algorithms for topic modelling. 
This has been made possible in other statistical 
domains by 
BUGS \citep[Bayesian inference using Gibbs sampling,][]{lunn2000winbugs} or 
JAGS \citep[just another Gibbs sampler,][]{plummer2003jags}, 
albeit with standard probability distributions. 
Theoretically, BUGS and JAGS will work on LDA; however, 
in practice, running Gibbs sampling for LDA with BUGS and JAGS 
is very slow.  This is because their Gibbs samplers 
are uncollapsed and not optimised. Furthermore, they
cannot be used in a model with stochastic processes, 
like the Gaussian process (GP) and DP.

Below, we present a framework that allows us to implement
HPYP topic models efficiently.
This framework allows us to test variants of our proposed topic models
without significant reimplementation.

\subsection{Hierarchical Pitman-Yor Process Topic Model}
\label{design_sec:generic_HPYP_topicmodel}

\begin{figure}[t!]
  \centering
  \includegraphics[width=0.7\linewidth]{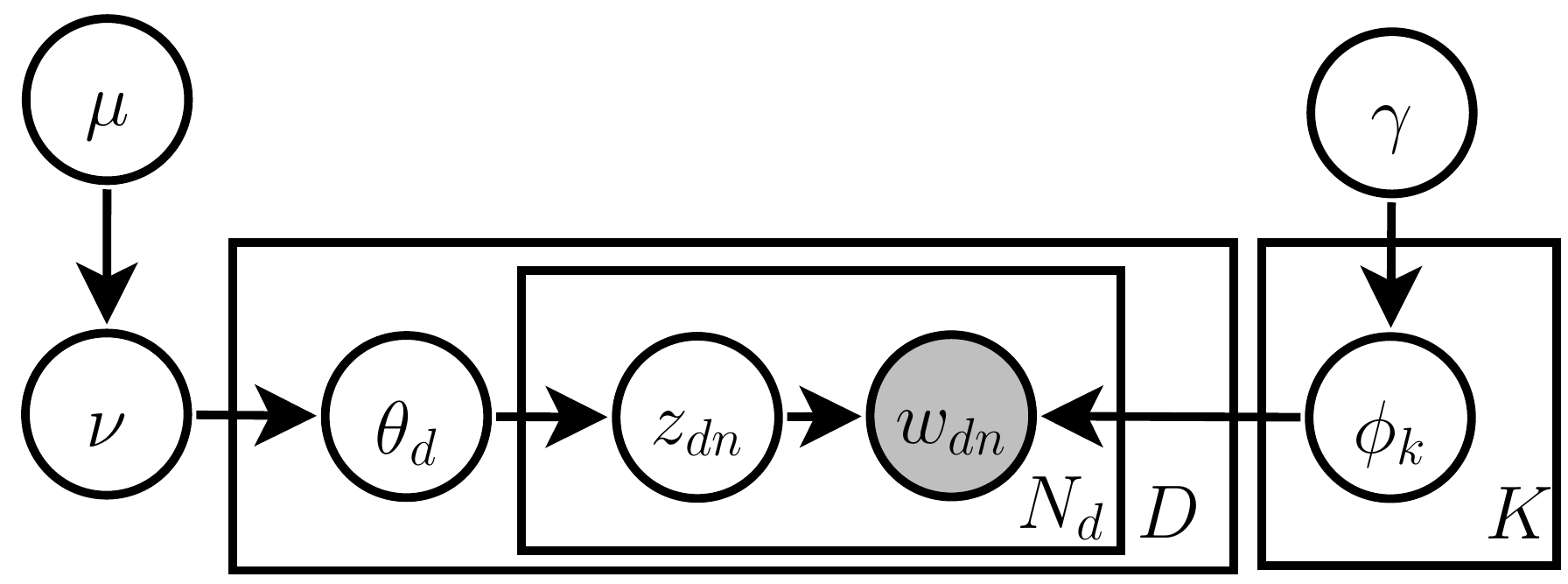}
  \vspace{-2mm}
  \caption[
      Graphical Model of the Hierarchical Pitman-Yor Process (HPYP) Topic Model
    ]{
      Graphical model of the HPYP topic model.
      It is an extension to LDA by allowing the probability
      vectors to be modelled by PYPs instead of the Dirichlet 
      distributions. 
      The area on the left of the graphical model (consists of 
      $\mu$, $\nu$ and $\theta$) is usually referred
      as topic side, while the right hand side (with $\gamma$
      and $\phi$) is called the vocabulary side.
      The word node denoted by $w_{dn}$ is observed.
      The notations are defined in Table~\ref{design_tbl:variables}.
    }
  \label{fig:generic_topicmodel}
\end{figure}

The HPYP topic model is a simple network of PYP 
nodes since all distributions on the probability vectors are modelled by the PYP. 
For simplicity, we assume a topic model with three PYP layers,
although in practice there is no limit to the number of PYP layers.
We present the graphical model of our generic topic model in Figure~\ref{fig:generic_topicmodel}.
This model is a variant of those presented in
\citet{Buntine:2014:ENT:2623330.2623691},
and is presented here as a starting model for illustrating
our methods and for subsequent extensions.

At the root level, we have $\mu$ and $\gamma$ distributed 
as PYPs:
\begin{align}
\mu  & \sim \mathrm{PYP}(\alpha^{\mu}, \beta^{\mu}, H^{\mu}) \,, \\
\gamma & \sim \mathrm{PYP}(\alpha^\gamma, \beta^\gamma, H^\gamma) \,.
\end{align}

\noindent
The variable $\mu$ is the root node for the \textit{topics} in a topic model
while $\gamma$ is the root node for the \textit{words}.
To allow arbitrary number of topics to be learned, we let the base distribution
for $\mu$, $H^{\mu}$, to be a continuous distribution or a discrete distribution
with infinite samples. 

We usually choose a discrete uniform distribution for $\gamma$ based on the 
word vocabulary size of the text corpus. 
This decision is technical in nature, as we are able to assign a tiny probability
to words not observed in the training set, which eases the evaluation process.
Thus $H^\gamma = \{ \cdots, \frac{1}{|\mathcal{V}|}, \cdots \}$ where $|\mathcal{V}|$ is the
set of all word vocabulary of the text corpus.

We now consider the topic side of the HPYP topic model.
Here we have $\nu$, which is the child node of $\mu$.
It follows a PYP given $\nu$, which acts as its base distribution:
\begin{align}
\nu \sim \mathrm{PYP}(\alpha^{\nu}, \beta^{\nu}, \mu) \,.
\end{align}

\noindent
For each document $d$ in a text corpus of size $D$, we have 
a document--topic distribution $\theta_d$\,, which is a topic distribution
specific to a document. Each of them tells us about the topic 
composition of a document.
\begin{align}
\theta_d \sim \mathrm{PYP}(\alpha^{\theta_d}, \beta^{\theta_d}, \nu) \,,
&& \mathrm{for\ \ } d = 1, \dots, D \,.
\end{align}

While for the vocabulary side, for each topic $k$ learned by the model,
we have a topic--word distribution $\phi_k$ which tells us about 
the words associated with each topic.
The topic--word distribution $\phi_k$ is PYP distributed given the parent node $\gamma$,
as follows:
\begin{align}
\phi_k  \sim \mathrm{PYP}(\alpha^{\phi_k}, \beta^{\phi_k}, \gamma) \,, 
&& \mathrm{for\ \ } k = 1, \dots, K \,.
\end{align}

\noindent
Here, $K$ is the number of topics in the topic model.

For every word $w_{dn}$ in a document $d$ which is indexed by $n$ 
(from $1$ to $N_d$\,, the number of words in document $d$),
we have a latent topic $z_{dn}$ (also known as topic assignment) which indicates the topic the word represents.
$z_{dn}$ and $w_{dn}$ are categorical variables generated from $\theta_d$ and 
$\phi_k$ respectively:
\begin{align}
z_{dn} \,|\, \theta_d     &\sim \mathrm{Discrete}(\theta_d)     \,, & 
\label{design_eq:topic}
\\
w_{dn} \,|\, z_{dn}, \phi &\sim \mathrm{Discrete}(\phi_{z_d})   \,, & 
\mathrm{for\ \ } n = 1, \dots, N_d \,.
\end{align}

\noindent
The above $\alpha$ and $\beta$ are the discount and concentration parameters of the PYPs 
(see Section~\ref{subsec:PYP}),
note that they are called the \textit{hyperparameters} in the model.
We present a list of variables used in this section in Table~\ref{design_tbl:variables}.

\begin{table}[t!]
    \centering
    \caption[
        List of Variables for the Hierarchical Pitman-Yor Process (HPYP) Topic Model
    ]{
        List of variables for the HPYP topic model used in this section.
    }
    \label{design_tbl:variables}
    \vspace{2mm}
    \begin{tabular}{cp{0.33\columnwidth}p{0.48\columnwidth}}
    \toprule
    \multicolumn{1}{c}{Variable} 
    & \multicolumn{1}{c}{Name}
    & \multicolumn{1}{c}{Description} 
    \\
    \midrule
    \multirow{1}{*}{$z_{dn}$}
    & \multirow{1}{\linewidth}{\centering Topic}
    & Topical label for word $w_{dn}$\,.
    \\
    \noalign{\vskip 2.5pt}
    \hdashline
    \noalign{\vskip 2.5pt}  
    \multirow{2}{*}{$w_{dn}$}
    & \multirow{2}{\linewidth}{\centering Word}
    & Observed word or phrase at position $n$ in document $d$.
    \\
    \noalign{\vskip 2.5pt}
    \hdashline
    \noalign{\vskip 2.5pt}  
    \multirow{2}{*}{$\phi_k$}
    & \multirow{2}{\linewidth}{\centering Topic--word distribution}
    & Probability distribution in generating words for topic $k$.
    \\
    \noalign{\vskip 2.5pt}
    \hdashline
    \noalign{\vskip 2.5pt}  
    \multirow{2}{*}{$\theta_d$}
    & \multirow{2}{\linewidth}{\centering Document--topic distribution}
    & Probability distribution in generating topics for document $d$.
    \\
    \noalign{\vskip 2.5pt} 
    \hdashline
    \noalign{\vskip 2.5pt}  
    \multirow{1}{*}{$\gamma$}
    & \centering Global word distribution 
    & \multirow{1}{\linewidth}{Word prior for $\phi_k$\,.}
    \\
    \noalign{\vskip 2.5pt} 
    \hdashline
    \noalign{\vskip 2.5pt}  
    \multirow{1}{*}{$\nu$} 
    & \centering Global topic distribution 
    & \multirow{1}{\linewidth}{Topic prior for $\theta_d$\,.}
    \\
    \noalign{\vskip 2.5pt} 
    \hdashline
    \noalign{\vskip 2.5pt}  
    \multirow{1}{*}{$\mu$} 
    & \centering Global topic distribution 
    & \multirow{1}{\linewidth}{Topic prior for $\nu$.}
    \\
    \noalign{\vskip 2.5pt} 
    \hdashline
    \noalign{\vskip 2.5pt}  
    $\alpha^\mathcal{N}$  
    & \centering Discount 
    & Discount parameter for PYP $\mathcal{N}$. 
    \\
    \noalign{\vskip 2.5pt} 
    \hdashline
    \noalign{\vskip 2.5pt} 
    \multirow{1}{*}{$\beta^\mathcal{N}$}
    & \multirow{1}{\linewidth}{\centering Concentration} 
    & Concentration parameter for PYP $\mathcal{N}$. 
    \\
    \noalign{\vskip 2.5pt} 
    \hdashline
    \noalign{\vskip 2.5pt}     
    $H^\mathcal{N}$ 
    & \centering Base distribution 
    & Base distribution for PYP $\mathcal{N}$. 
    \\
    \noalign{\vskip 2.5pt} 
    \hdashline
    \noalign{\vskip 2.5pt}     
    \multirow{2}{*}{$c^\mathcal{N}_k$}
    & \multirow{2}{\linewidth}{\centering Customer count}
    & Number of customers having dish $k$ in restaurant $\mathcal{N}$. 
    \\ 
    \noalign{\vskip 2.5pt} 
    \hdashline
    \noalign{\vskip 2.5pt}     
    \multirow{2}{*}{$t^\mathcal{N}_k$}
    & \multirow{2}{\linewidth}{\centering Table count}
    & Number of tables serving dish $k$ in restaurant $\mathcal{N}$. 
    \\ 
    \noalign{\vskip 2.5pt} 
    \hdashline
    \noalign{\vskip 2.5pt}     
    $\mathbf{Z}$ 
    & \centering All topics 
    & Collection of all topics $z_{dn}$\,. 
    \\
    \noalign{\vskip 2.5pt} 
    \hdashline
    \noalign{\vskip 2.5pt}     
    $\mathbf{W}$ 
    & \centering All words 
    & Collection of all words $w_{dn}$\,. 
    \\
    \noalign{\vskip 2.5pt} 
    \hdashline
    \noalign{\vskip 2.5pt}     
    \multirow{2}{*}{$\mathbf{\Xi}$}
    & \multirow{2}{\linewidth}{\centering All hyperparameters}
    & Collection of all hyperparameters and constants.
    \\
    \noalign{\vskip 2.5pt} 
    \hdashline
    \noalign{\vskip 2.5pt}     
    $\mathbf{C}$ 
    & \centering All customer counts
    & Collection of all customers counts $c^\mathcal{N}_k$. 
    \\
    \noalign{\vskip 2.5pt} 
    \hdashline
    \noalign{\vskip 2.5pt}     
    $\mathbf{T}$ 
    & \centering All table counts
    & Collection of all table counts $t^\mathcal{N}_k$. 
    \\
    \bottomrule
    \end{tabular}
\end{table}

\subsection{Model Representation and Posterior Likelihood}
\label{design_sec:model_representation}

In a Bayesian setting, posterior inference requires us
to analyse the posterior distribution of the model variables given the observed data.
For instance, the joint posterior distribution for the HPYP topic model is
\begin{align}
p(\mu, \nu, \gamma, \theta, \phi, \mathbf{Z} \,|\, \mathbf{W}, \mathbf{\Xi}) \,.
\label{design_eq:original_joint_posterior}
\end{align}

\noindent
Here, we use bold face capital letters 
to represent the set of all relevant 
variables. 
For instance, $\mathbf{W}$ captures all words in the corpus. 
Additionally, we denote $\mathbf{\Xi}$ as the set of 
all hyperparameters and constants in the model. 

Note that deriving the posterior distribution analytically is almost impossible
due to its complex nature.
This leaves us with approximate Bayesian inference techniques as mentioned in 
Section~\ref{subsec:mcmc}.
However, even with these techniques, performing posterior inference with 
the posterior distribution is difficult due to the coupling of the probability
vectors from the PYPs.

\begin{figure}[t!]
    \centering
    \includegraphics[width=0.95\linewidth]{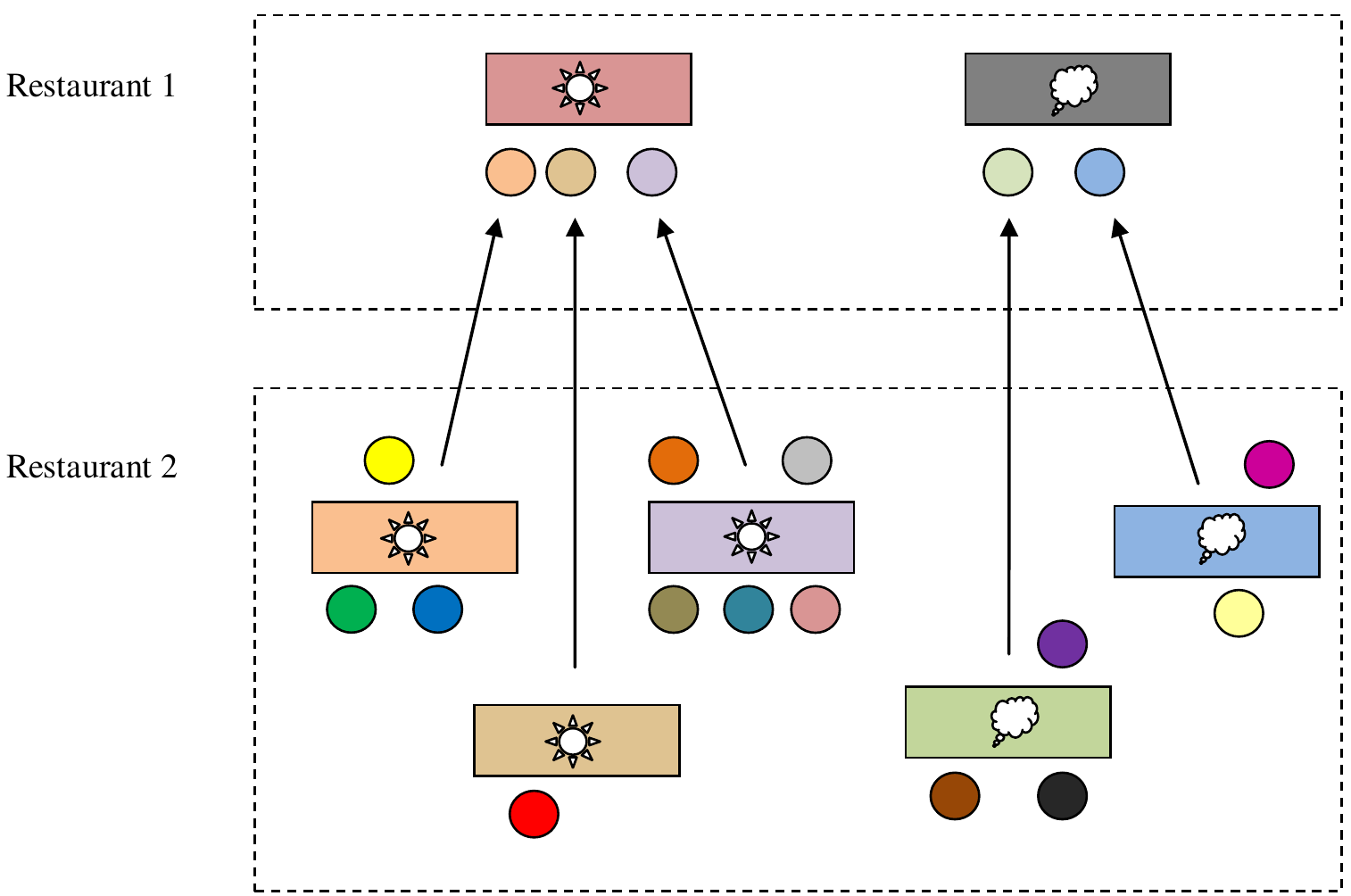}
    \vspace{-2mm}
    \caption[
        Illustration of the Chinese Restaurant Representation
    ]{
        An illustration of the Chinese restaurant process representation. 
        The customers are represented by the circles while the tables
        are represented by the rectangles.
        The dishes are the symbols in the middle of the
        rectangles, here they are denoted by the sunny symbol
        and the cloudy symbol.
        In this illustration, we know the number of customers corresponds
        to each table, for example, the green table is occupied by
        three customers.
        Also, since Restaurant 1 is the parent of Restaurant 2, the tables
        in Restaurant 2 are treated as the customers for Restaurant 1.        
    }
    \label{fig:chinese_restaurant1}
\end{figure}

The key to an efficient inference procedure with the PYPs is to marginalise out 
the PYPs in the model and record various associated counts instead, which yields a 
collapsed sampler.
To achieve this, we adopt a Chinese Restaurant Process (CRP) metaphor 
\citep{TehJor2010a, Blei:2010:NCR:1667053.1667056} to represent the variables in the topic model. 
With this metaphor, all data in the model ({\it e.g.}, topics and words) are the {\it customers};
while the PYP nodes are the {\it restaurants} the customers visit.
In each restaurant, each customer is to be seated at only one {\it table}, 
though each table can have \textit{any} number of customers.
Each table in a restaurant serves a {\it dish}, the dish corresponds to the 
categorical label a data point may have ({\it e.g.}, the topic label or word).
Note that there can be more than one table serving the same dish.
In a HPYP topic model, the tables in a restaurant $\mathcal{N}$ are treated as 
the customers for the parent restaurant $\mathcal{P}$ 
(in the graphical model, $\mathcal{P}$ points to $\mathcal{N}$), 
and they share the same dish.
This means that the data is passed up recursively until the root node.
For illustration, we present a simple example in Figure~\ref{fig:chinese_restaurant1},
showing the seating arrangement of the customers from two restaurants.

\begin{figure}[t!]
    \centering
    \includegraphics[width=0.95\linewidth]{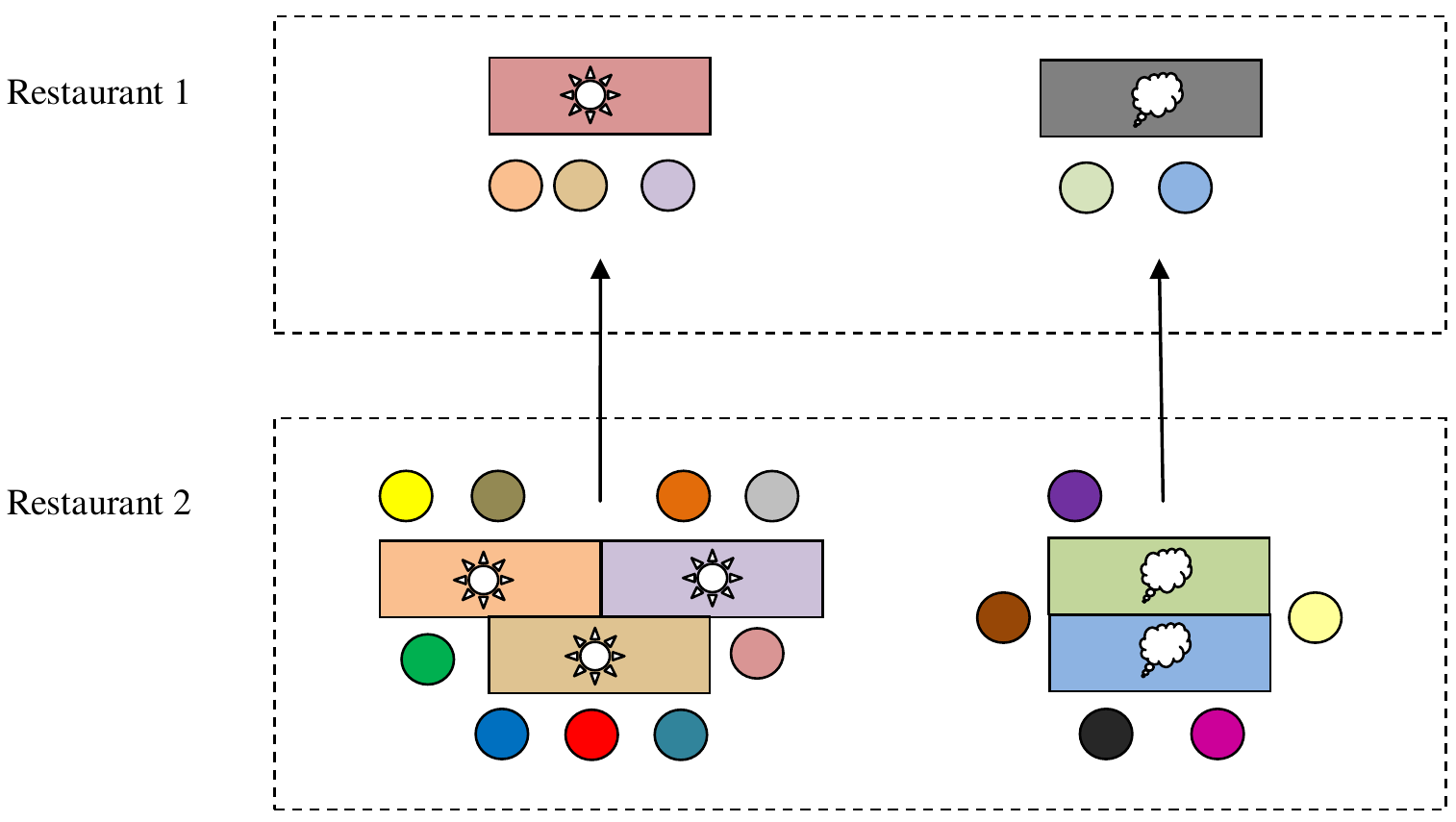}
    \vspace{-2mm}
    \caption[
        Illustration of the Chinese Restaurant with the Table Multiplicity Representation
    ]{
        An illustration of the Chinese restaurant with the table counts representation.
        Here the setting is the same as Figure~\ref{fig:chinese_restaurant1} but the 
        seating arrangement of the customers are ``forgotten'' and 
        only the table and customer counts are recorded. Thus, 
        we only know that there are three sunny tables in Restaurant 2, with a total of
        nine customers.
    }
    \label{fig:chinese_restaurant2}
\end{figure}

Na\"ively recording the seating arrangement (table and dish) of each customer brings about 
computational inefficiency during inference. 
Instead, we adopt the table multiplicity (or table counts) representation 
of \citet{Chen:2011:STC:2034063.2034095} which requires no dynamic memory, 
thus consuming only a factor of memory at no loss of inference 
efficiency.
Under this representation, we store only the customer counts and table counts 
associated with each restaurant.
The customer count $c^\mathcal{N}_k$ denotes the number of customers who are having
dish $k$ in restaurant $\mathcal{N}$. The corresponding symbol without subscript, 
$c^\mathcal{N}$, denotes the collection of customer counts in restaurant $\mathcal{N}$,
that is, $c^\mathcal{N} = (\cdots, c^\mathcal{N}_k, \cdots)$.
The total number of customers in a restaurant $\mathcal{N}$ 
is denoted by the capitalised symbol instead, $C^\mathcal{N} = \sum_k c^\mathcal{N}_k$.
Similar to the customer count, the table count $t^\mathcal{N}_k$ denotes 
the number of non-empty tables serving dish $k$ in restaurant $\mathcal{N}$.
The corresponding $t^\mathcal{N}$ and $T^\mathcal{N}$ are defined similarly.
For instance, from the example in Figure~\ref{fig:chinese_restaurant1}, we have
$c^2_{\mathrm{sun}} = 9$ and $t^2_{\mathrm{sun}} = 3$, the corresponding illustration
of the table multiplicity representation is presented in Figure~\ref{fig:chinese_restaurant2}.
We refer the readers to \citet{Chen:2011:STC:2034063.2034095}
for a detailed derivation of the posterior likelihood of 
a restaurant.

For the posterior likelihood of the HPYP topic model, 
we marginalise out the probability vector associated with the PYPs 
and represent them with the customer counts
and table counts, following \citet[][Theorem~1]{Chen:2011:STC:2034063.2034095}.
We present the modularised version of the full posterior of 
the HPYP topic model, 
which allows the posterior to be computed very quickly.
The full posterior consists of the modularised 
likelihood associated with each PYP in the model, defined as
\begin{align}
f(\mathcal{N}) & = 
\frac{
  \big( \beta^\mathcal{N} \big| \alpha^\mathcal{N} \big)_{{T}^\mathcal{N}} 
}{
  \big( \beta^\mathcal{N} \big)_{C^\mathcal{N}}
} 
\prod_{k=1}^K 
S^{c^\mathcal{N}_k}_{{t}^\mathcal{N}_k,\, \alpha^\mathcal{N}} 
\binom{c^\mathcal{N}_k}{t^\mathcal{N}_k}^{-1}
\,,
&&
\mathrm{for\ \ }
\mathcal{N} \sim 
\mathrm{PYP} \big( \alpha^\mathcal{N}, \beta^\mathcal{N}, \mathcal{P} \big) 
\,.  
\label{design_eq:modularized_likelihood} 
\end{align}%

\noindent
Here,
$S^x_{y,\,\alpha}$ are generalised Stirling numbers
\citep[Theorem~17]{Buntine:2010arXiv1007.0296B}. 
Both $(x)_T$ 
and $(x|y)_T$ denote Pochhammer symbols with rising factorials
\citep[Section~18]{Oldham:2008}:
\begin{align}
(x)_T   & = x\cdot(x+1) \cdots \big(x+(T-1)\big) ~\,,    \\
(x|y)_T & = x\cdot(x+y) \cdots \big(x+(T-1)y\big) \,.
\end{align}

\noindent
With the CRP representation, the full posterior of the 
HPYP topic model can now be written 
--- 
in terms of $f(\cdot)$ given in 
Equation~\eqref{design_eq:modularized_likelihood} --- as 
\begin{align}
p(\mathbf{Z}, \mathbf{T}, \mathbf{C} \,|\, \mathbf{W}, \mathbf{\Xi}) & \propto 
p(\mathbf{Z}, \mathbf{W}, \mathbf{T}, \mathbf{C} \,|\, \mathbf{\Xi})  \nonumber \\[1pt]
& \propto
f(\mu) f(\nu) 
\Bigg( \prod_{d=1}^D f(\theta_d) \! \Bigg)
\Bigg( \prod_{k=1}^K f(\phi_k) \! \Bigg)
f(\gamma) 
\Bigg( 
\prod_{v=1}^{|\mathcal{V}|} \left(\frac{1}{|\mathcal{V}|}\right)^{t^\gamma_v} 
\Bigg) .
\label{design_eq:likelihood}
\end{align}

\noindent
This result is a generalisation of \citet[Theorem~1]{Chen:2011:STC:2034063.2034095}
to account for discrete base distribution ---
the last term in Equation~\eqref{design_eq:likelihood} corresponds 
to the base distribution of $\gamma$, and $v$ indexes each unique word 
in vocabulary set $\mathcal{V}$.
The bold face $\mathbf{T}$ and $\mathbf{C}$ denote the collection of all
table counts and customer counts, respectively.
Note that the topic assignments $\mathbf{Z}$ are implicitly captured 
by the customer counts:
\begin{align}
c^{\theta_d}_k = \sum_{n=1}^{N_d} I(z_{dn} = k)  \,,
\label{design_eq:counts_for_topics}
\end{align}

\noindent
where $I(\cdot)$ is the indicator function, which evaluates to $1$ 
when the statement inside the function is true, and $0$ otherwise.
We would like to point out that 
even though the probability vectors of the PYPs are integrated out and 
not explicitly stored, they can easily be reconstructed. 
This is discussed in Section~\ref{subsec:estimate_pyp}.
We move on to Bayesian inference in the next section.

\section{Posterior Inference for the HPYP Topic Model}
\label{design_sec:posterior_inference}

We focus on the MCMC method for 
Bayesian inference on the HPYP topic model.
The MCMC method on topic models follows these simple procedures ---
decrementing counts contributed by a word, 
sample a new topic for the word, and 
update the model by accepting or rejecting the proposed sample.
Here, we describe the collapsed blocked Gibbs sampler for the HPYP topic model.
Note the PYPs are marginalised out so we only deal with the~counts.

\subsection{Decrementing the Counts Associated with a Word}
\label{subsec:decrement}

The first step in a Gibbs sampler is to remove a word and corresponding
latent topic, then decrement the associated customer counts and table
counts.
To give an example from Figure~\ref{fig:chinese_restaurant1}, 
if we remove the red customer from Restaurant 2, we would decrement
the customer count $c^2_{\mathrm{sun}}$ by $1$. 
Additionally, we also decrement the table
count $t^2_{\mathrm{sun}}$ by $1$ because the red customer is 
the only customer on its table.
This in turn decrements the customer count $c^1_{\mathrm{sun}}$ by $1$.
However, this requires us to keep track of the customers'
seating arrangement which leads to increased memory requirements and poorer performance
due to inadequate mixing \citep{Chen:2011:STC:2034063.2034095}.

To overcome the above issue, we follow the concept of table indicator 
\citep{Chen:2011:STC:2034063.2034095} and introduce a new auxiliary
Bernoulli indicator variable $u^\mathcal{N}_k$, which indicates
whether removing the customer also removes the table by which the customer is
seated.
Note that our Bernoulli indicator is different to that of
\citet{Chen:2011:STC:2034063.2034095} which indicates the 
restaurant a customer contributes to.
The Bernoulli indicator is sampled as needed in the decrementing 
procedure and it is not stored, this means that we simply ``forget''
the seating arrangements and re-sample them later when needed,
thus we do not need to store the seating arrangement.
The Bernoulli indicator of a restaurant $\mathcal{N}$ 
depends solely on the customer counts and the table counts:
\begin{align}
p \big( u^\mathcal{N}_k \big) =
\left\{
    \begin{array}{ll}
        t^{\mathcal{N}}_k / c^\mathcal{N}_k  & ~~\mathrm{if ~~} u^\mathcal{N}_k = 1 \\[6pt]
        1 - t^{\mathcal{N}}_k / c^\mathcal{N}_k & ~~\mathrm{if ~~} u^\mathcal{N}_k = 0
        ~\,.
    \end{array}
\right. 
\label{eq:bernoulli_indicator}
\end{align}

In the context of the HPYP topic model described in 
Section~\ref{design_sec:generic_HPYP_topicmodel},
we formally present how we decrement the counts associated
with the word $w_{dn}$ and latent topic $z_{dn}$ 
from document $d$ and position $n$.
First, on the vocabulary side (see Figure~\ref{fig:generic_topicmodel}), 
we decrement the 
customer count $c^{\phi_{z_{dn}}}_{w_{dn}}$ 
associated with $\phi_{z_{dn}}$ by 1.
Then sample a Bernoulli indicator $u^{\phi_{z_{dn}}}_{w_{dn}}$
according to Equation~\eqref{eq:bernoulli_indicator}.
If $u^{\phi_{z_{dn}}}_{w_{dn}} = 1$, we decrement the table
count $t^{\phi_{z_{dn}}}_{w_{dn}}$ and also the customer count
$c^\gamma_{w_{dn}}$ by one. In this case, we would sample 
a Bernoulli indicator $u^\gamma_{w_{dn}}$ for $\gamma$, 
and decrement $t^\gamma_{w_{dn}}$ if $u^\gamma_{w_{dn}} = 1$.
We do not decrement the respective customer count if the Bernoulli
indicator is $0$.
Second, we would need to decrement the counts associated with
the latent topic $z_{dn}$\,. The procedure is similar, we decrement
$c^{\theta_d}_{z_{dn}}$ by $1$ and sample the Bernoulli
indicator $u^{\theta_d}_{z_{dn}}$\,.
Note that whenever we decrement a customer count, we sample the
corresponding Bernoulli indicator.
We repeat this procedure recursively until the Bernoulli indicator
is $0$ or until the procedure hits the root node.


\subsection{Sampling a New Topic for a Word}
\label{subsec:block_sampling}

After decrementing the variables associated with a word $w_{dn}$\,, 
we use a {\it blocked} Gibbs sampler to sample a new topic $z_{dn}$ for the word 
and the corresponding customer counts and table counts.
The conditional posterior used in sampling can be computed quickly
when the full posterior is represented in a modularised form.
To illustrate, the conditional posterior for $z_{dn}$ and its associated 
customer counts and table counts~is
\begin{align}
p(z_{dn}, \mathbf{T}, \mathbf{C} \,|\, \mathbf{Z}^{-dn}, \mathbf{W}, \mathbf{T}^{-dn}, \mathbf{C}^{-dn}, \mathbf{\Xi}) 
= \frac{ p(\mathbf{Z}, \mathbf{T}, \mathbf{C} \,|\, \mathbf{W}, \mathbf{\Xi}) 
}{ p(\mathbf{Z}^{-dn}, \mathbf{T}^{-dn}, \mathbf{C}^{-dn} \,|\, \mathbf{W}, \mathbf{\Xi}) } \,,
\label{design_eq:conditional_posterior}
\end{align}

\noindent
which is further broken down by substituting the posterior likelihood 
defined in Equation~\eqref{design_eq:likelihood}, giving the following ratios of 
the modularised likelihoods:
\begin{align}
\frac{ f(\mu)         }{ f(\mu^{-dn})         }
\frac{ f(\nu)         }{ f(\nu^{-dn})         }
\frac{ f(\theta_d)      }{ f(\theta_d^{-dn})      } 
\frac{ f(\phi_{z_{dn}}) }{ f(\phi_{z_{dn}}^{-dn}) } 
\frac{ f(\gamma)        }{ f(\gamma^{-dn})        }
\left(\frac{1}{|\mathcal{V}|}\right)^{t^\gamma_{w_{dn}} - {(t^\gamma_{w_{dn}})}^{-dn}} 
\,.
\label{design_eq:broken_modularised_likelihood}
\end{align}

\noindent
The superscript $\Box^{-dn}$ indicates that the variables associated 
with the word $w_{dn}$ are removed from the respective sets,
that is, the customer counts and table counts are after the decrementing procedure.
Since we are only sample the topic assignment $z_{dn}$ associated
with one word, the customer counts and table counts can only increment by at most $1$,
see Table~\ref{design_tbl:proposals} for a list of all possible proposals.
%

\begin{table}[t!]
    \centering
    \caption[
        All Possible Proposals of the Blocked Gibbs Sampler
    ]{
        All possible proposals of the blocked Gibbs sampler for the variables associated
        with $w_{dn}$\,. To illustrate, one sample would be $z_{dn}=1$, 
        $t^\mathcal{N}_{z_{dn}}$ does not increment (stays the same),
        and $c^\mathcal{N}_{z_{dn}}$ increments by 1,
        for all $\mathcal{N}$ in $\{ \mu, \nu, \theta_d, \phi_{z_{dn}}, \gamma \}$.
        We note that the proposals can include states that are invalid, but this is
        not an issue since those states have zero posterior probability and 
        thus will not be sampled.
    }
    \label{design_tbl:proposals}
    \begin{tabular}{cccccc}
    \toprule
    Variable 
    & Possibilities
    & Variable
    & Possibilities
    & Variable
    & Possibilities
    \\
    \midrule
    $z_{dn}$
    & $\{1, \dots, K\}$
    & $t^\mathcal{N}_{z_{dn}}$
    & $\{t^\mathcal{N}_{z_{dn}},\, t^\mathcal{N}_{z_{dn}} + 1\}$
    & $c^\mathcal{N}_{z_{dn}}$
    & $\{c^\mathcal{N}_{z_{dn}},\, c^\mathcal{N}_{z_{dn}} + 1\}$
    \\
    \bottomrule
    \end{tabular}
\end{table}
This allows the ratios of the modularised likelihoods,
which consists of ratios of Pochhammer symbol and ratio of Stirling numbers
\begin{align}
\frac{ f(\mathcal{N}) }{ f({\mathcal{N}}^{-dn}) } =
\frac{ (\beta^\mathcal{N})_{{(C^\mathcal{N})}^{-dn}}
}{ (\beta^\mathcal{N})_{C^\mathcal{N}} } 
\frac{ (\beta^\mathcal{N}|\alpha^\mathcal{N})_{{T}^\mathcal{N}} 
}{ (\beta^\mathcal{N}|\alpha^\mathcal{N})_{{(T^\mathcal{N})}^{-dn}} } 
\prod_{k=1}^K 
\frac{ S^{c^\mathcal{N}_k}_{{t}^\mathcal{N}_k,\, \alpha^\mathcal{N}} 
}{ S^{{(c^\mathcal{N}_k)}^{-dn}}_{{(t^\mathcal{N}_k)}^{-dn},\, \alpha^\mathcal{N}} }
~\,,
\label{design_eq:likelihood_ratio}
\end{align}

\noindent
to simplify further.
For instance, the ratios of Pochhammer symbols can be reduced to constants, as follows:
\begin{align}
\frac{ (x)_{T+1}   }{ (x)_T   } = x + T    \,, &&
\frac{ (x|y)_{T+1} }{ (x|y)_T } = x + yT   \,.
\end{align}

\noindent
The ratio of Stirling numbers, such as 
$S^{y+1}_{x+1,\, \alpha}/S^{y}_{x,\, \alpha}$\,,
can be computed quickly via caching \citep{Buntine:2010arXiv1007.0296B}.
Technical details on implementing the Stirling numbers cache
can be found in \citet{LimThesis}.

With the conditional posterior defined, we proceed to the sampling process.
Our first step involves finding all possible changes to the
topic $z_{dn}$\,, customer counts,
and the table counts (hereafter known as `{\it state}') 
associated with adding the removed word $w_{dn}$ back
into the topic model.
Since only one word is added into the model, the customer counts and
the table counts can only increase by at most $1$,
constraining the possible states to a reasonably small number.
Furthermore, the customer counts of a parent node will only be
incremented when the table counts of its child node increases.
Note that it is possible for the added customer to generate
a new dish (topic) for the model.
This requires the customer to increment the table count of a \textit{new} dish in
the root node $\mu$ by $1$ (from $0$).

Next, we compute the conditional posterior 
(Equation~\eqref{design_eq:conditional_posterior})
for all possible states.
The conditional posterior (up to a proportional constant) 
can be computed quickly by breaking down the posterior and calculating
the relevant parts.
We then normalise them to sample one of the states to be 
the proposed next state.
Note that the proposed state will always be accepted,
which is an artifact of Gibbs sampler.

Finally, given the proposal, we update the HPYP model by
incrementing the relevant customer counts and table counts.


\subsection{Optimising the Hyperparameters}
\label{subsec:hyperparameter}

Choosing the right hyperparameters for the
priors is important for topic models.
\citet{WallachPrior2009} show that an 
optimised hyperparameter increases
the robustness of the topic models and improves
their model fitting.
The hyperparameters of the HPYP topic models 
are the discount parameters and 
concentration parameters
of the PYPs.
Here, we propose a procedure to optimise the concentration parameters,
but leave the discount parameters fixed due to their coupling 
with the Stirling numbers cache.

The concentration parameters $\beta$ of all the PYPs 
are optimised using an auxiliary variable sampler 
similar to \citet{Teh06abayesian}.
Being Bayesian, we assume the concentration parameter $\beta^\mathcal{N}$
of a PYP node $\mathcal{N}$ has the following {\it hyperprior}:
\begin{align}
\beta^\mathcal{N} \sim \mathrm{Gamma}(\tau_0, \tau_1) \,,
&&
\mathrm{for\ \ }
\mathcal{N} \sim 
\mathrm{PYP}\big(\alpha^\mathcal{N}, \beta^\mathcal{N}, \mathcal{P} \big) 
\,, 
\end{align}

\noindent
where $\tau_0$ is the {\it shape} parameter and $\tau_1$ is the {\it rate} parameter.
The gamma prior is chosen due to its conjugacy 
which gives a gamma posterior for $\beta^\mathcal{N}$.

To optimise $\beta^\mathcal{N}$, we first 
sample the auxiliary variables $\omega$ and $\zeta_i$ 
given the \textit{current} value
of $\alpha^\mathcal{N}$ and $\beta^\mathcal{N}$, as follows:
\begin{align}
\omega \,|\, \beta^\mathcal{N} 
& \sim \mathrm{Beta} \big( C^\mathcal{N}, \beta^\mathcal{N} \big)
\,, \\[1pt]
\zeta_i \,|\, \alpha^\mathcal{N}, \beta^\mathcal{N} &\sim
\mathrm{Bernoulli}\left(\frac{\beta^\mathcal{N}}{\beta^\mathcal{N}+i\alpha^\mathcal{N}}\right) \,,
&& \mathrm{for\ \ } i = 0, 1, \dots, T^\mathcal{N} - 1 ~\,.
\end{align}

\noindent
With these, we can then sample a new $\beta^\mathcal{N}$ 
from its conditional posterior
\begin{align}
\beta^\mathcal{N} \,\big|\, \omega, \zeta \sim 
\mathrm{Gamma}\left(\tau_0 +  \sum_{i=0}^{T^\mathcal{N}-1}  \zeta_i ~~ , ~~
\tau_1 - \log(1 - \omega)\right) ~\,.
\end{align}
%
%
The collapsed Gibbs sampler is summarised by
Algorithm~\ref{alg:Gibbs_sampler}.

\begin{algorithm}[t!]
\caption{Collapsed Gibbs Sampler for the HPYP Topic Model}
\label{alg:Gibbs_sampler}
\begin{enumerate}
\vspace{2mm}
\item 
    Initialise the HPYP topic model by assigning random topic 
    to the latent topic $z_{dn}$ associated to each word $w_{dn}$\,. 
    Then update all the relevant customer counts $\mathbf{C}$ 
    and table counts $\mathbf{T}$
    by using Equation~\eqref{design_eq:counts_for_topics} and setting
    the table counts to be about half of the customer counts.
\item 
    For each word $w_{dn}$ in each document $d$, do the following:
    \begin{enumerate}[label=(\alph*)]
        \item 
            Decrement the counts associated with $w_{dn}$ 
            (see Section~\ref{subsec:decrement}).
        
       \item 
           Block sample a new topic for $z_{dn}$ and corresponding
           customer counts $\mathbf{C}$ and table counts $\mathbf{T}$ 
           (see Section~\ref{subsec:block_sampling}).
       \item
           Update (increment counts) the topic model based on the sample.
       \end{enumerate}
\item 
    Update the hyperparameter $\beta^\mathcal{N}$ for each PYP nodes $\mathcal{N}$
    (see Section~\ref{subsec:hyperparameter}).
\item 
    Repeat Steps 2\,--\,3 until the model converges or when a fix number of iterations is~reached.
\end{enumerate}
\end{algorithm}

\subsection{Estimating the Probability Vectors of the PYPs}
\label{subsec:estimate_pyp}

Recall that the aim of topic modelling is to analyse the posterior 
of the model parameters, such as one in Equation~\eqref{design_eq:original_joint_posterior}.
Although we have marginalised out the PYPs in the above
Gibbs sampler, the PYPs can be reconstructed from the associated
customer counts and table counts.
Recovering the full posterior distribution of the PYPs
is a complicated task. So, instead, we will analyse 
the PYPs {\it via} the expected value
of their conditional marginal posterior distribution,
or simply, their {\it posterior mean},
\begin{align}
\mathbb{E}[\mathcal{N}\,|\,\mathbf{Z},\mathbf{W},\mathbf{T},\mathbf{C},\mathbf{\Xi}] \,, &&
\mathrm{for\ } \mathcal{N} \in \{ \mu, \nu, \gamma, \theta_d, \phi_k \} \,.
\end{align}

The posterior mean of a PYP
corresponds to the probability of sampling a new customer for the PYP.
To illustrate, we consider the posterior of the topic distribution $\theta_d$\,.
We let $\tilde{z}_{dn}$ to be a unknown {\it future} latent topic in
addition to the known $\mathbf{Z}$.
With this, we can write the posterior mean of~$\theta_{dk}$~as
\begin{align}
\mathbb{E}[\theta_{dk}\,|\,\mathbf{Z},\mathbf{W},\mathbf{T},\mathbf{C},\mathbf{\Xi}]
& = \mathbb{E}[ p(\tilde{z}_{dn} = k \,|\, \theta_d,
\mathbf{Z},\mathbf{W},\mathbf{T},\mathbf{C},\mathbf{\Xi}) \,|\, 
\mathbf{Z},\mathbf{W},\mathbf{T},\mathbf{C},\mathbf{\Xi}]
\nonumber \\[1pt]
& = \mathbb{E}[ p(\tilde{z}_{dn} = k \,|\, \mathbf{Z}, \mathbf{T}, \mathbf{C}) \,|\, 
\mathbf{Z},\mathbf{W},\mathbf{T},\mathbf{C},\mathbf{\Xi}]
\,.
\end{align}

\noindent
by replacing $\theta_{dk}$ with the posterior predictive distribution
of $\tilde{z}_{dn}$ and note that $\tilde{z}_{dn}$ can be sampled
using the CRP, as follows:
\begin{align}
p(\tilde{z}_{dn} = k \,|\, \mathbf{Z}, \mathbf{T}, \mathbf{C})
= 
\frac{
(\alpha^{\theta_d} T^{\theta_d} + \beta^{\theta_d}) \nu_k 
+ c_k^{\theta_d} - \alpha^{\theta_d} T_k^{\theta_d}
}{
\beta^{\theta_d} + C^{\theta_d} 
}  \,.
\end{align}

\noindent
Thus, the posterior mean of $\theta_d$ 
is given as
\begin{align}
\mathbb{E}[\theta_{dk}\,|\,\mathbf{Z},\mathbf{W},\mathbf{T},\mathbf{C},\mathbf{\Xi}]
= 
\frac{
(\alpha^{\theta_d} T^{\theta_d} + \beta^{\theta_d}) 
\mathbb{E}[\nu_k\,|\,\mathbf{Z},\mathbf{W},\mathbf{T},\mathbf{C},\mathbf{\Xi}]
+ c_k^{\theta_d} - \alpha^{\theta_d} T_k^{\theta_d}
}{
\beta^{\theta_d} + C^{\theta_d} 
}
\,,
\end{align}

\noindent
which is written in term of the posterior mean of its parent PYP, $\nu$.
The posterior means of the other PYPs such as $\nu$ can be derived 
by taking a similar approach.
Generally, the posterior mean corresponds to a PYP $\mathcal{N}$ 
(with parent PYP $\mathcal{P}$)
is as follows:
\begin{align}
\mathbb{E}[\mathcal{N}_k \,|\, \mathbf{Z},\mathbf{W},\mathbf{T},\mathbf{C},\mathbf{\Xi}]
= 
\frac{
(\alpha^\mathcal{N} T^\mathcal{N} + \beta^\mathcal{N}) 
\mathbb{E}[\mathcal{P}_k\,|\,\mathbf{Z},\mathbf{W},\mathbf{T},\mathbf{C},\mathbf{\Xi}]
+ c_k^\mathcal{N} - \alpha^\mathcal{N} T_k^\mathcal{N}
}{
\beta^\mathcal{N} + C^\mathcal{N}
}
\,,
\label{design_eq:recover_vector}
\end{align}

\noindent
By applying Equation~\eqref{design_eq:recover_vector} recursively,
we obtain the posterior mean for all the PYPs in the model.

We note that the dimension of the topic distributions
($\mu$, $\nu$, $\theta$)
is $K+1$, where $K$ is the number of observed topics.
This accounts for the generation of a new topic 
associated with the new customer, though the probability
of generating a new topic is usually much smaller. 
In practice, we may instead ignore the extra dimension
during the evaluation of a topic model since it does not
provide useful interpretation.
One way to do this is to simply discard the extra dimension
of all the probability vectors after computing the posterior mean.
Another approach would be to normalise the posterior mean of 
the root node $\mu$ after discarding the extra dimension,
before computing the posterior mean of others PYPs.
Note that for a considerably large corpus, the difference
in the above approaches would be too small to notice.

\subsection{Evaluations on Topic Models}

Generally, there are two ways to evaluate a topic model. 
The first is to evaluate the topic model based on the task
it performs, for instance, 
the ability to make predictions.
The second approach is the statistical evaluation
of the topic model on modelling the data, which is also
known as the goodness-of-fit test.
In this section, we will present some commonly used
evaluation metrics that are applicable to all topic models,
but we first discuss the procedure for estimating
variables associated with the test set.

\subsubsection{Predictive Inference on the Test Documents}
\label{subsec:predictive_inference}

Test documents, which are used for evaluations, are set aside 
from learning documents. As such, the document--topic distributions 
${\theta}$
associated with the test documents are unknown and hence
need to be estimated.
One estimate for ${\theta}$ is its posterior mean 
given the variables learned from the Gibbs sampler:
\begin{align}
\hat{{\theta}}_d 
= \mathbb{E}[{\theta}_d \,|\, \mathbf{Z}, \mathbf{W}, \mathbf{T}, \mathbf{C}, \mathbf{\Xi}] \,,
\label{design_eq:test_doc_distribution}
\end{align}

\noindent
obtainable by applying Equation~\eqref{design_eq:recover_vector}.
Note that since the latent topics $\tilde{\mathbf{Z}}$ 
corresponding to the test set are not sampled, 
the customer counts and table counts associated with 
${\theta}_d$ are $0$, thus $\hat{{\theta}}_d$
is equal to $\hat{\nu}$, the posterior mean of $\nu$.
However, this is not a good estimate for the topic 
distribution of the test documents since they 
will be identical for all the test documents.
To overcome this issue, we will instead use some of the words
in the test documents to obtain a better estimate for ${\theta}$.
This method is known as document completion \citep{Wallach:2009:EMT:1553374.1553515},
as we use part of the text to estimate ${\theta}$,
and use the rest for evaluation. 

Getting a better estimate for ${\theta}$ requires us to
first sample some of the latent topics $\tilde{z}_{dn}$ in the test documents.
The proper way to do this is by running an algorithm akin 
to the collapsed Gibbs sampler, but this would be excruciatingly slow
due to the need to re-sample the customer counts and table counts 
for all the parent PYPs.
Instead, we assume that the variables learned from the Gibbs sampler
are fixed and sample the $\tilde{z}_{dn}$ 
from their conditional posterior sequentially,
given the previous latent topics:
\begin{align}
p(\tilde{z}_{dn} = k 
\,|\, \tilde{w}_{dn}, {\theta}_d, \phi, \tilde{z}_{d1}, \dots, \tilde{z}_{d,n-1} )
\propto {\theta}_{dk} \, \phi_{kw_{dn}}
\,.
\end{align}

\noindent
Whenever a latent topic $\tilde{z}_{dn}$ is sampled, we increment
the customer count $c^{{\theta}_d}_{\tilde{z}_{dn}}$ for 
the test document.
For simplicity, we set the table count $t^{{\theta}_d}_{\tilde{z}_{dn}}$ 
to be half the corresponding customer counts $c^{{\theta}_d}_{\tilde{z}_{dn}}$, 
this avoids the expensive operation of sampling the table counts. 
Additionally, ${\theta}_d$ is re-estimated using 
Equation~\eqref{design_eq:test_doc_distribution} 
before sampling the next latent topic.
We note that the estimated variables are unbiased.

The final ${\theta}_d$ becomes an estimate 
for the topic distribution of the test document $d$.
The above procedure is repeated $R$ times to give
$R$ samples of ${\theta}_d^{(r)}$, which are used
to compute the following Monte Carlo estimate of ${\theta}_d$:
\begin{align}
\hat{{\theta}}_d
= \frac{1}{R} \sum_{r=1}^R {\theta}_d^{(r)}
\,.
\end{align}

\noindent
This Monte Carlo estimate can then be used for
computing the evaluation metrics.
Note that when estimating ${\theta}$,
we have ignored the possibility of generating
a new topic, that is, the latent topics $\tilde{z}$ 
are constrained to the existing topics,
as previously discussed in Section~\ref{subsec:estimate_pyp}.

\subsubsection{Goodness-of-fit Test}
\label{subsec:goodness-of-fit}

Measures of goodness-of-fit usually involves computing 
the discrepancy of the observed values and 
the predicted values under the model.
However, the observed variables in a topic model
are the words in the corpus, which are not quantifiable
since they are discrete labels.
Thus evaluations on topic models are usually
based on the model likelihoods instead.

A popular metric commonly used to evaluate the goodness-of-fit 
of a topic model is perplexity, which is negatively related to the likelihood 
of the observed words $\mathbf{W}$ given the model, this is defined as
\begin{align}
\mathrm{perplexity}(\mathbf{W}\,|\, \theta, \phi) 
= 
\exp\Bigg( -
\frac{
    \sum_{d=1}^D \sum_{n=1}^{N_d} \log p(w_{dn} \,|\, \theta_d, \phi)
}{
    \sum_{d=1}^D N_d
}
\Bigg) \,,
\end{align}

\noindent
where $p(w_{dn}\,|\,\theta_d, \phi)$ is 
the likelihood of sampling the word $w_{dn}$
given the document--topic distribution $\theta_d$ and
the topic--word distributions $\phi$.
Computing $p(w_{dn}\,|\,\theta_d, \phi)$
requires us to marginalise out $z_{dn}$
from their joint distribution, as follows:
\begin{align}
p(w_{dn} \,|\, \theta_d, \phi) 
& = \sum_k p(w_{dn}, \, z_{dn} = k \,|\, \theta_d, \phi) 
\nonumber \\
& = \sum_k p(w_{dn}\,|\,z_{dn}=k, \phi_k) \, p(z_{dn}=k\,|\,\theta_d) 
\nonumber \\
& = \sum_k \phi_{kw_{dn}} \theta_{dk} ~\,.
\label{design_eq:word_probability}
\end{align}

Although perplexity can be computed on the whole corpus,
in practice we compute the perplexity on test documents. 
This is to measure if the topic model generalises 
well to unseen data.
A good topic model would be able to predict the 
words in the test set better, thereby assigning a higher
probability $p(w_{dn}\,|\,\theta_d, \phi)$ in generating the words.
Since perplexity is negatively related to the likelihood,
a lower perplexity is better.

\subsubsection{Document Clustering}
\label{subsubsec:clustering}

We can also evaluate the clustering ability of the topic models.
Note that topic models assign a topic to each word in a document, 
essentially performing a \textit{soft clustering} 
\citep{EroshevaFienbergWeihsGaul2005} for the documents
in which the membership is 
given by the document--topic distribution~$\theta$.
To evaluate the clustering of the documents, 
we convert the soft clustering to hard 
clustering by choosing a topic that best represents the documents, 
hereafter called the \textit{dominant topic}. The dominant topic of
a document $d$
corresponds to the topic that has the highest proportion in the topic 
distribution, that is,
\begin{align}
\mathrm{Dominant\ Topic}(\theta_d) = \argmax_k \theta_{dk}
\,.
\end{align}

Two commonly used evaluation measures for clustering are 
\textit{purity} and \textit{normalised mutual information}
\citep[NMI,][]{Manning:2008:IIR:1394399}.
The purity is a simple clustering measure which can be interpreted as 
the proportion of documents correctly clustered,
while NMI is an information theoretic measures used for clustering
comparison.
If we denote the
ground truth classes as $\mathcal{S} = \{s_1, \dots, s_J\}$ and 
the obtained clusters as $\mathcal{R} = \{r_1, \dots, r_K\}$,
where each $s_j$ and $r_k$ represents a collection (set) of documents,
then the purity 
and NMI can be computed as 
\begin{align}
\mathrm{purity}(\mathcal{S}, \mathcal{R}) = 
\frac{1}{D} \sum_{k=1}^K \max_j | r_k \cap s_j | \,,
&& 
\mathrm{NMI}(\mathcal{S}, \mathcal{R}) = 
\frac{2 \, \mathrm{MI}(\mathcal{S}; 
\mathcal{R})}{E(\mathcal{S})+E(\mathcal{R})} \,,
\end{align}

\noindent
where $\mathrm{MI}(\mathcal{S}; \mathcal{R})$ denotes the mutual information 
between two sets
and $E(\cdot)$ denotes the entropy. They are defined as follows:
\begin{align}
\mathrm{MI}(\mathcal{S}; \mathcal{R}) = 
\sum_{k=1}^K \sum_{j=1}^J 
\frac{| r_k \cap s_j |}{D} 
\log_2 D \frac{ |r_k \cap s_j|}{| r_k | | s_j |} \,,
&& 
E(\mathcal{R}) = - \sum_{k=1}^K \frac{|r_k|}{D} \log_2 \frac{|r_k|}{D} 
\,.
\end{align}
Note that the higher the purity or NMI, the better the clustering.

\section{Application: Modelling Social Network on Twitter}
\label{sec:application}

This section looks at how we can employ the framework discussed
above for an application of tweet modelling, 
using auxiliary information 
that is available on Twitter. 
We propose the {\em Twitter-Network topic model} (TNTM) to 
jointly model the text and the social network in a fully Bayesian 
nonparametric way,
in particular, by incorporating the authors, hashtags, the
``follower'' network, and the text content in modelling.
The TNTM employs a HPYP
for text modelling and a Gaussian process (GP) random function model for social network 
modelling. We show that the TNTM significantly outperforms several existing 
nonparametric models due to its flexibility.

\subsection{Motivation}
\label{tntm_sec:introduction}

Emergence of web services such as blogs, microblogs and social networking websites allows 
people to contribute information freely and publicly. This user-generated information is 
generally more personal, informal, and often contains personal opinions. In 
aggregate, it can be useful for 
reputation analysis of entities and products \citep{Aula2010}, 
natural disaster detection \citep{Karimi:2013:CMD:2537734.2537737}, 
obtaining first-hand news \citep{BroersmaGraham2012}, 
or even demographic analysis \citep{Correa2010}. 
We focus on Twitter, an
accessible source of information that 
allows users to freely voice their opinions and thoughts 
in short text known as tweets.

Although LDA \citep{Blei:2003:LDA:944919.944937} is a 
popular model for text modelling,
a direct application on tweets often yields poor result as tweets are 
short and often noisy \citep{Zhao:2011:CTT:1996889.1996934, baldwin-EtAl:2013:IJCNLP}, 
that is, 
tweets are unstructured and often contain grammatical and spelling errors, as well as 
{\emph{informal}} words such as user-defined abbreviations due to the 140 characters 
limit. LDA fails on short tweets since it is heavily dependent on word co-occurrence. Also 
notable is that the text in tweets may contain special tokens known as {\em{hashtags}}; 
they are used as keywords and allow users to link their tweets with other tweets tagged 
with the same hashtag. Nevertheless, hashtags are informal since they have no standards. 
Hashtags can be used as both inline words or categorical labels. When used as labels, 
hashtags are often noisy, since users can create new hashtags easily and use any existing 
hashtags in any way they like.%
\footnote{
    For example, {\em hashtag hijacking}, where a well defined 
    hashtag is used in an ``inappropriate'' way. The most notable example
    would be on the hashtag \textit{\#McDStories}, though it was initially created to
    promote happy stories on McDonald's, the hashtag was hijacked with negative
    stories on McDonald's.
} 
Hence instead of being hard labels, hashtags are best treated as special words 
which can be the themes of the tweets.
These properties of tweets make them challenging for topic models, 
and {\it ad hoc} alternatives are used instead. 
For instance, \cite{maynard2012challenges} advocate the use of shallow method for 
tweets, and \cite{Mehrotra:2013:ILT:2484028.2484166} utilise a tweet-pooling 
approach to group short 
tweets into a larger document.
In other text analysis applications, tweets are often `cleansed' by NLP methods such 
as lexical normalisation \citep{baldwin-EtAl:2013:IJCNLP}. However, the use of normalisation 
is also criticised \citep{Eisenstein:2013}, as normalisation can change the meaning of text.

In the following, we propose a novel method for better modelling of microblogs by leveraging 
the auxiliary information that accompanies tweets. This information, complementing word 
co-occurrence, also opens the door to more 
applications, such as user recommendation and hashtag suggestion. Our major contributions 
include
(1)~a fully Bayesian nonparametric model named the Twitter-Network topic model (TNTM) 
that models tweets well, and
(2)~a combination of both the HPYP and the GP to jointly model text, hashtags, authors 
and the followers network.
Despite the seeming complexity of the TNTM model, its implementation is made relatively 
straightforward using the flexible framework developed in Section~\ref{sec:framework}. Indeed, a number of other variants were rapidly implemented
with this framework as well.

\subsection{The Twitter-Network Topic Model}
\label{tntm_sec:model}

\begin{figure}[tb!]
    \centering
    \includegraphics[width=\linewidth]{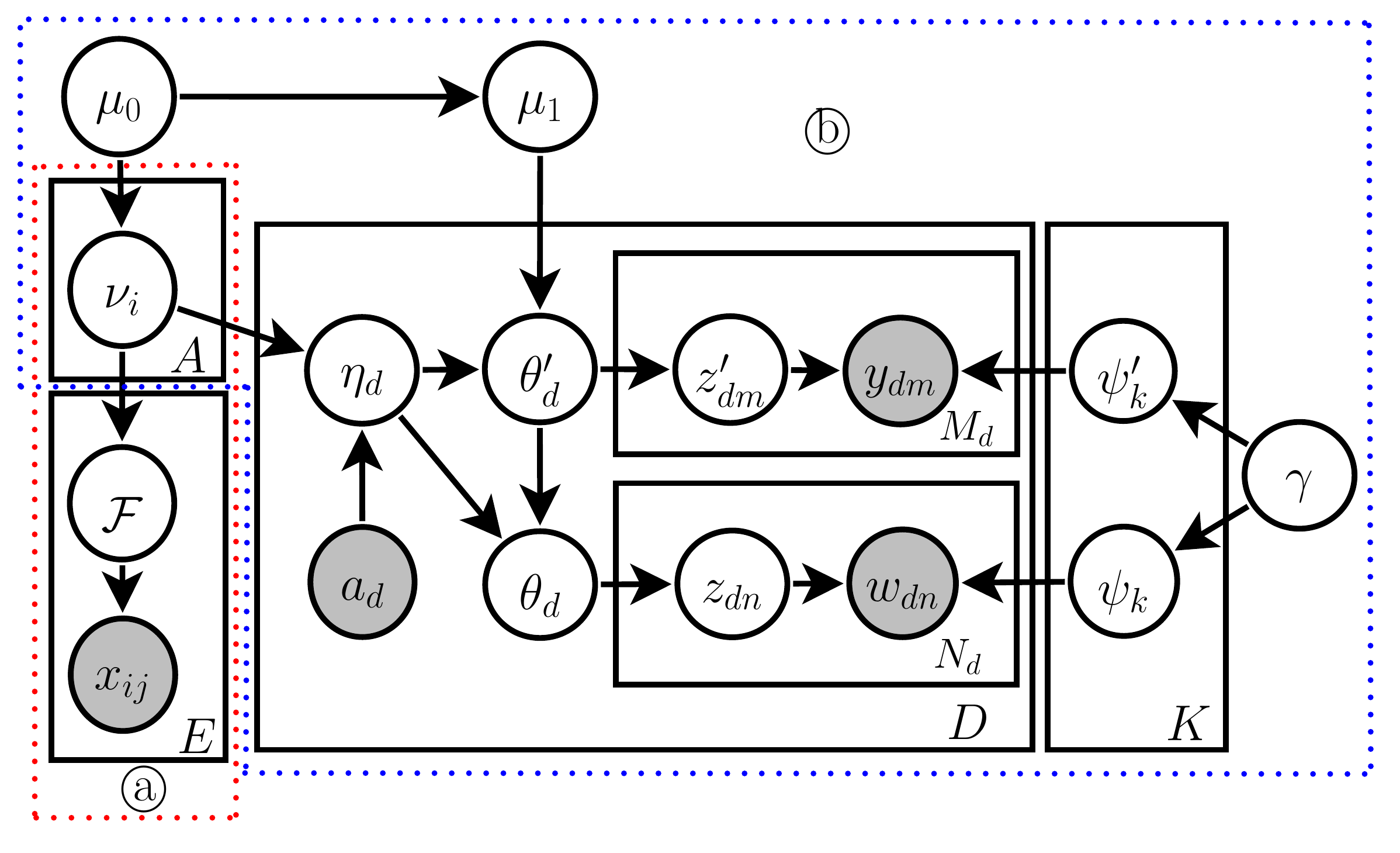}
    \vspace{-6mm}
    \caption[
        Graphical Model for the Twitter-Network Topic Model (TNTM)
    ]{
          Graphical model for the Twitter-Network Topic Model (TNTM)
          composed of a HPYP topic model (region~\textcircled{b})
          and a GP based random function network model (region~\textcircled{a}).
          The author--topic distributions $\nu$ serve to link the two
          together. Each tweet is modelled with a hierarchy of document--topic
          distributions denoted by $\eta$, $\theta'$, and $\theta$, where
          each is attuned to the whole tweet, the hashtags, and the words,
          in that order. With their own topic assignments $z'$ and $z$, 
          the hashtags $y$ and the words $w$ are separately modelled.
          They are generated from the topic--hashtag distributions $\psi'$ and 
          the topic--word distributions $\psi$ respectively. The variables
          $\mu_0$\,, $\mu_1$ and $\gamma$ are priors for the respective PYPs.
          The connections between the authors are denoted by $x$, 
          modelled by random function~$\mathcal{F}$.
    }
    \label{tntm_fig:twitter_network}
\end{figure}

The TNTM makes use of the accompanying {\em hashtags}, {\em authors}, and 
{\em followers network} to model tweets better. The TNTM is composed of two main 
components: a HPYP topic model for the text and hashtags, and a GP based random function 
network model for the followers network. The authorship information serves to connect the two 
together. The HPYP topic model is illustrated by region~\textcircled{b} in
Figure~\ref{tntm_fig:twitter_network} while the network model is captured
by region~\textcircled{a}.

\subsubsection{HPYP Topic Model}

The HPYP topic model described in Section~\ref{sec:framework} is extended as follows. 
For the word distributions, we first 
generate a parent word distribution prior $\gamma$ for all topics:
\begin{align}
\gamma \sim \mathrm{PYP}(\alpha^\gamma, \beta^\gamma, H^\gamma) \,,
\end{align}

\noindent
where $H_\gamma$ is a discrete uniform distribution over the complete word vocabulary 
$\mathcal{V}$.%
\footnote{
    The complete word vocabulary contains words and hashtags seen in the corpus.
}
Then, we sample the hashtag distribution $\psi'_k$ and word distribution $\psi_k$ for 
each topic~$k$, with $\gamma$ as the base distribution:
\begin{align}
\psi'_k \, | \, \gamma & \sim \mathrm{PYP}(\alpha^{\psi'_k}, \beta^{\psi'_k}, \gamma) \,, 
\\
\psi_k  \, | \, \gamma & \sim \mathrm{PYP}(\alpha^{\psi_k}, \beta^{\psi_k}, \gamma) \,,
&& \mathrm{for\ \ } k = 1, \dots, K \,.
\end{align}

\noindent
Note that the tokens of the hashtags are shared with the words, that is, the hashtag 
\textit{\#happy} shares the same token as the word \textit{happy}, and are thus
treated as the same word. This treatment is important 
since some hashtags are used as words instead of labels.%
\footnote{
    For instance, as illustrated by the following tweet:
    \textit{
        i want to get into \#photography. can someone recommend a good beginner \#camera please? 
        i dont know where to start
    }
}
Additionally, this also allows any 
words to be hashtags, which will be useful for hashtag recommendation.

For the topic distributions, we generate a global topic distribution $\mu_0$,
which serves as a prior, from a GEM distribution.
Then generate the author--topic distribution $\nu_i$ for each author $i$, and a 
miscellaneous topic distribution $\mu_1$ to capture topics that deviate from the authors' 
usual topics:
\begin{align}
\mu_0 & \sim \mathrm{GEM}(\alpha^{\mu_0}, \beta^{\mu_0}) \,, \\
\mu_1 \,|\, \mu_0 & \sim \mathrm{PYP}(\alpha^{\mu_1}, \beta^{\mu_1}, \mu_0) \,, \\
\nu_i \,|\, \mu_0 & \sim \mathrm{PYP}(\alpha^{\nu_i}, \beta^{\nu_i}, \mu_0) \,,
&& \mathrm{for\ \ } i = 1, \dots, A \,.
\end{align}

\noindent
For each tweet $d$, given the author--topic distribution $\nu$ and the observed author $a_d$\,, 
we sample the document--topic distribution 
$\eta_d$\,, as follows:
\begin{align}
\eta_d \,|\, a_d, \nu \sim \mathrm{PYP}(\alpha^{\eta_d}, \beta^{\eta_d}, \nu_{a_d}) \,,
&& \mathrm{for\ \ } d = 1, \dots, D \,.
\end{align}

\noindent
Next, we generate the topic distributions for the observed hashtags ($\theta'_d$) and the 
observed words ($\theta_d$), following the technique used in the adaptive topic model 
\citep{Du:2012:MST:2390948.2391010}.
We explicitly model the influence of hashtags to words, by generating the words 
conditioned on the hashtags. The intuition comes from hashtags being the themes of a 
tweet, and they drive the content of the tweet. Specifically, we sample the mixing 
proportions $\rho^{\theta'_d}$, which control the contribution of $\eta_d$ and $\mu_1$ for 
the base distribution of $\theta'_d$\,, and then generate $\theta'_d$ given $\rho^{\theta'_d}$:
\begin{align}
\rho^{\theta'_d} & \sim \mathrm{Beta}\Big(\lambda^{\theta'_d}_0, \lambda^{\theta'_d}_1 \Big) 
\,, \\
\theta'_d \,|\, \mu_1, \eta_d 
& \sim \mathrm{PYP}\Big(\alpha^{\theta'_d}, \beta^{\theta'_d}, 
\rho^{\theta'_d} \mu_1 + (1\!-\!\rho^{\theta'_d}) \eta_d \Big) \,.
\end{align}

\noindent
We set $\theta'_d$ and $\eta_d$ as the parent distributions of $\theta_d$\,. This 
flexible configuration allows us to investigate the relationship between $\theta_d$\,, 
$\theta'_d$ and $\eta_d$\,, that is, we can examine if $\theta_d$ is directly determined 
by $\eta_d$\,, or through the $\theta'_d$\,. The mixing proportions $\rho^{\theta_d}$ and 
the topic distribution $\theta_d$ is generated similarly:
\begin{align}
\rho^{\theta_d} & \sim \mathrm{Beta}\Big(\lambda^{\theta_d}_0, \lambda^{\theta_d}_1 \Big) 
\,, \\
\theta_d \,|\, \eta_d, \theta'_d 
& \sim \mathrm{PYP}\Big(\alpha^{\theta_d}, \beta^{\theta_d}, \,
\rho^{\theta_d} \eta_m + (1\!-\!\rho^{\theta_d}) \theta'_d \Big) \,.
\end{align}

\noindent
The hashtags and words are then generated in a similar fashion to LDA. 
For the $m$-th hashtag in tweet $d$, we sample a topic $z'_{dm}$ and the hashtag $y_{dm}$ by
\begin{align}
z'_{dm} \,|\, \theta'_d & \sim \mathrm{Discrete}\big( \theta'_d \big) \,, \\
y_{dm}  \,|\, z'_{dm}, \psi' & \sim \mathrm{Discrete}\Big( \psi'_{z'_{dm}} \Big) \,,
&& \mathrm{for\ \ } m = 1, \dots, M_d \,,
\end{align}

\noindent
where $M_d$ is the number of seen hashtags in tweet $d$.
While for the $n$-th word in tweet $d$, we sample a topic $z_{dn}$ and the word $w_{dn}$ by
\begin{align}
z_{dn} \,|\, \theta_d & \sim \mathrm{Discrete}(\theta_d) \,, \\ 
w_{dn} \,|\, z_{dn}, \psi & \sim \mathrm{Discrete}\big( \psi_{z_{dn}} \big) \,,
&& \mathrm{for\ \ } n = 1, \dots, N_d \,,
\end{align}

\noindent
where $N_d$ is the number of observed words in tweet $d$.
We note that all above $\alpha$, $\beta$ and $\lambda$ are the hyperparameters of 
the model.
We show the importance of the above modelling with ablation studies in 
Section~\ref{tntm_sec:experiment}.
Although the HPYP topic model may seem complex, it is a simple network of PYP 
nodes since all distributions on the probability vectors are modelled by the PYP. 

\subsubsection{Random Function Network Model}

The network modelling is connected to the HPYP topic model {\it{via}} the author--topic 
distributions $\nu$, where we treat $\nu$ as inputs to the GP in the network model. The 
GP, represented by $\mathcal{F}$, determines the link between two authors ($x_{ij}$), which 
indicates the existence of the social links between author $i$ and author $j$. 
For each pair of authors, we sample their connections with the following random 
function network model:
\begin{align}
Q_{ij} \,|\, \nu & \sim \mathcal{F}(\nu_i, \nu_j) \,, \\
x_{ij} \,|\, Q_{ij} & \sim \mathrm{Bernoulli}\big( s(Q_{ij}) \big) \,,
&& \mathrm{for\ \ } i = 1, \dots, A; \ j = 1, \dots, A \,,
\end{align}
where $s(\cdot)$ is the \textit{sigmoid function}:
\begin{align}
s(t) = \frac{1}{1 + e^{-t}} ~.
\end{align}

\noindent
By marginalising out $\mathcal{F}$, we can write
$\mathbf{Q} \sim \mathrm{GP}(\varsigma, \kappa)$,
where $\mathbf{Q}$ is a \textit{vectorised} collection of $Q_{ij}$\,.%
\footnote{
    $\mathbf{Q} = (Q_{11}, Q_{12}, \dots, Q_{AA})^\mathsf{T} $,
    note that $\varsigma$ and $\kappa$ follow the same indexing.
}
$\varsigma$ denotes the mean vector and $\kappa$ is 
the covariance matrix of the GP:
\begin{align}
\varsigma_{ij} & = \mathrm{Sim}(\nu_i, \nu_j) \,, 
\label{tntm_eq:mean_function}
\\
\kappa_{ij, i'j'} & =
\frac{s^2}{2} \exp \! 
  \Bigg( \! 
    -\frac{
      \big| \mathrm{Sim}(\nu_{i}, \nu_{j}) - \mathrm{Sim}(\nu_{i'}, \nu_{j'}) \big|^2
    }{
      2l^2
    }
  \Bigg) 
  + \sigma^2 I(ij = i'j') \,, 
\label{tntm_eq:kernel}
\end{align}

\noindent
where $s$, $l$ and $\sigma$ are the hyperparameters associated to the kernel. 
$\mathrm{Sim}(\cdot, \cdot)$ is a similarity function that has a range
between $0$ and $1$, here chosen to be 
{\em cosine similarity} due to its
ease of computation and popularity.

\subsubsection{Relationships with Other Models}

The TNTM is related to many existing models after removing certain components of the model.
When hashtags and the network components are removed, the TNTM is reduced to a 
nonparametric variant of the author topic model (ATM). 
Oppositely, if authorship information is discarded, the TNTM resembles the 
{\em correspondence LDA} \citep{Blei:2003:MAD:860435.860460}, although it differs 
in that it allows hashtags and words to be generated from a common vocabulary.

In contrast to existing parametric models, 
the network model in the TNTM provides possibly the most flexible way of network modelling 
{\em via} a nonparametric Bayesian prior~(GP), following \citet{lloyd2012random}. 
Different to \cite{lloyd2012random}, 
we propose a new kernel function that fits our purpose better and achieves significant 
improvement over the original~kernel. 
  
\subsection{Representation and Model Likelihood}
\label{tntm_sec:representation}

As with previous sections, we represent the TNTM using the CRP representation
discussed in Section~\ref{design_sec:model_representation}.
However, since the PYP variables in the TNTM can have multiple parents,
we extend the representation following \cite{Du:2012:MST:2390948.2391010}.
The distinction is that we store multiple tables counts for each PYP, 
to illustrate, $t_k^{\mathcal{N} \to \mathcal{P}}$ represents the
number of tables in PYP $\mathcal{N}$ serving dish $k$ that are 
contributed to the customer counts in PYP $\mathcal{P}$, $c_k^\mathcal{P}$.
Similarly, the total table counts that contribute to $\mathcal{P}$ is
denoted as $T^{\mathcal{N} \to \mathcal{P}} = \sum_k t_k^{\mathcal{N} \to \mathcal{P}}$.
Note the number of tables in PYP $\mathcal{N}$ is 
$t_k^\mathcal{N} = \sum_\mathcal{P} t_k^{\mathcal{N} \to \mathcal{P}}$, 
while the total number of tables is 
$T^\mathcal{N} = \sum_\mathcal{P} T^{\mathcal{N} \to \mathcal{P}}$.
We refer the readers to \citet[Appendix~B]{Lim2013Twitter}
for a detailed~discussion.

We use bold face capital letters to denote the set of all relevant lower case variables,
for example, we denote $\mathbf{W}^\circ = \{\mathbf{W}, \mathbf{Y}\}$ as the set of all words and hashtags; 
$\mathbf{Z}^\circ = \{\mathbf{Z}, \mathbf{Z}'\}$ as the set of all 
topic assignments for the words and the hashtags; 
$\mathbf{T}$ as the set of all table counts and $\mathbf{C}$ as the set of all customer counts; 
and we introduce $\mathbf{\Xi}$ as the set of all hyperparameters.
By marginalising out the latent variables, we write down the model likelihood 
corresponding to the HPYP topic model in terms of the counts:
\begin{align}
p(\mathbf{Z}^\circ, \mathbf{T}, \mathbf{C} \, | \, \mathbf{W}^\circ, \mathbf{\Xi})
\propto & \
p(\mathbf{Z}^\circ, \mathbf{W}^\circ, \mathbf{T}, \mathbf{C} \, | \, \mathbf{\Xi}) 
\nonumber
\\
\propto & \
f(\mu_0)f(\mu_1) \Bigg( \prod_{i=1}^A f(\nu_i) \Bigg) 
\Bigg( \prod_{k=1}^K f(\psi'_k)f(\psi_k) \Bigg) f(\gamma) 
\nonumber
\\
& \ \ \ 
\times \Bigg( 
  \prod_{d=1}^D f(\eta_d) f(\theta'_d) f(\theta_d) 
  g \big( \rho^{\theta'_d} \big) g \big(\rho^\theta_d \big)
\Bigg) 
\prod_{v=1}^{|\mathcal{V}|} \bigg( \frac{1}{|\mathcal{V}|} \bigg)^{t_v^\gamma}
,
\label{tntm_eq:hypy_posterior} 
\end{align}

\noindent
where $f(\mathcal{N})$ is the modularised likelihood corresponding to node $\mathcal{N}$,
as defined by Equation~\eqref{design_eq:modularized_likelihood}, and 
$g(\rho)$ is the likelihood corresponding to the probability $\rho$ that controls which 
parent node to send a customer to, defined~as
\begin{align}
g(\rho^\mathcal{N}) & = 
B \Big(
  \lambda^\mathcal{N}_{0} + T^{\mathcal{N} \to \mathcal{P}_0}, \
  \lambda^\mathcal{N}_{1} + T^{\mathcal{N} \to \mathcal{P}_1}
\Big) 
\,,
\end{align}

\noindent
for $\mathcal{N} \sim 
\mathrm{PYP} \big( 
  \alpha^\mathcal{N}, 
  \beta^\mathcal{N},  \,
  \rho^\mathcal{N} \mathcal{P}_0 + (1\!-\!\rho^\mathcal{N}) \mathcal{P}_1  
\big)$.
Note that 
$B(a, b)$ denotes the Beta function 
that normalises a Dirichlet distribution, defined as follows:
\begin{align}
B(a,b) = \frac{\Gamma(a) \, \Gamma(b)}{\Gamma(a+b)}
~.
\end{align}

%

\noindent
For the random function network model, the conditional posterior can be derived as 
\begin{align}
p(\mathbf{Q} \, | \, \mathbf{X}, \nu, \mathbf{\Xi}) 
\propto & \
p(\mathbf{X}, \mathbf{Q} \, | \, \nu, \mathbf{\Xi}) 
\nonumber
\\[1pt]
\propto & \
\Bigg( \prod_{i=1}^A \prod_{j=1}^A 
  s(Q_{ij})^{x_{ij}} \,
  \Big( 1 - s(Q_{ij}) \Big)^{1 - x_{ij}} 
\Bigg)
\nonumber
\\
& \ \ \ \ \
\times 
|\kappa|^{-\frac{1}{2}} \,
\exp \! \bigg( \!
  -\frac{1}{2} (\mathbf{Q} - \varsigma)^\mathsf{T} \, \kappa^{-1} \, (\mathbf{Q} - \varsigma)
\bigg)
\,.
\label{tntm_eq:network_posterior}
\end{align}

\noindent
The full posterior likelihood is thus the product of the topic model posterior 
(Equation~\eqref{tntm_eq:hypy_posterior}) and the network posterior  
(Equation~\eqref{tntm_eq:network_posterior}):
\begin{align}
p(
  \mathbf{Q}, \mathbf{Z}^\circ, \mathbf{T}, \mathbf{C} \, | \,
  \mathbf{X}, \mathbf{W}^\circ, \mathbf{\Xi}
)
=
p(\mathbf{Z}^\circ, \mathbf{T}, \mathbf{C} \, | \, \mathbf{W}^\circ, \mathbf{\Xi})
\,
p(\mathbf{Q} \, | \, \mathbf{X}, \nu, \mathbf{\Xi}) 
\,.
\end{align}

\subsection{Performing Posterior Inference on the TNTM}
\label{tntm_sec:inference}

In the TNTM, combining a GP with a HPYP makes its posterior inference non-trivial. 
Hence, we employ approximate inference by alternatively performing MCMC sampling 
on the HPYP topic model and the network model, conditioned on each other. 
For the HPYP topic model, we 
employ the flexible framework discussed in Section~\ref{sec:framework} 
to perform collapsed blocked Gibbs sampling.
For the network model, we derive a Metropolis-Hastings (MH) algorithm based 
on the elliptical slice sampler \citep{murray2009elliptical}. In addition, the 
author--topic distributions $\nu$ connecting the HPYP and the GP are sampled with an MH scheme
since their posteriors do not follow a standard form.
We note that the PYPs in this section can have multiple parents,
so we extend the framework in Section~\ref{sec:framework} to allow for this.


The collapsed Gibbs sampling for the HPYP topic model in TNTM is similar to
the procedure in Section~\ref{design_sec:posterior_inference}, although there are
two main differences.
The first difference is that we need to sample the topics for both words and hashtags, 
each with a different conditional posterior compared to that of 
Section~\ref{design_sec:posterior_inference}.
While the second is due to the PYPs in TNTM can have multiple parents,
thus an alternative to decrementing the counts is required.
A detailed discussion on performing posterior inference
and hyperparameter sampling is 
presented in the appendix.%

\subsection{Twitter Data}
\label{tntm_sec:data}

For evaluation of the TNTM, we construct a tweet corpus from the {\em Twitter 7} 
dataset \citep{Yang:2011:PTV:1935826.1935863},%
\footnote{
    \url{http://snap.stanford.edu/data/twitter7.html}
}
This corpus is queried using the hashtags \textit{\#sport}, \textit{\#music}, \textit{\#finance},
\textit{\#politics}, \textit{\#science} and \textit{\#tech}, chosen for diversity.
We remove the non-English tweets with {\it langid.py} \citep{Lui:2012:LOL:2390470.2390475}.
We obtain the data on the followers network from \cite{Kwak:2010:TSN:1772690.1772751}.%
\footnote{
    \url{http://an.kaist.ac.kr/traces/WWW2010.html}
}
However, note that this followers network data is not complete and does not
contain information for all authors.
Thus we filter out the authors that are not part of the followers network data
from the tweet corpus.
Additionally, we also remove authors who have written less than fifty tweets
from the corpus.
We name this corpus T6 since it is queried with six hashtags. 
It is consists of 240,517 tweets with 150 authors after filtering.

Besides the T6 corpus, we also use the tweet datasets
described in \cite{Mehrotra:2013:ILT:2484028.2484166}. 
The datasets contains three corpora, each of them is queried
with exactly ten query terms. The first corpus, named the 
Generic Dataset, are queried with generic terms.
The second is named the Specific Dataset, which is composed 
of tweets on specific named entities.
Lastly, the Events Dataset is associated with certain events.
The datasets are mainly used for comparing the performance of the
TNTM against the tweet pooling techniques in \cite{Mehrotra:2013:ILT:2484028.2484166}.
We present a summary of the tweet corpora in 
Table~\ref{tntm_tbl:datasets}.


\begin{table}[t!]
    \centering
    \caption[
        Summary of the Datasets
    ]{
        Summary of the datasets used in this section, showing the 
        number of tweets ($D$), authors ($A$), unique word 
        tokens ($|\mathcal{V}|$), and
        the average number of words and hashtags in each tweet.
        The T6 dataset is queried with six different hashtags
        and thus has a higher number of hashtags per tweet.
        We note that there is a typo on the number of tweets for
        the Events Dataset in \cite{Mehrotra:2013:ILT:2484028.2484166},
        the correct number is 107,128.
    }
    \label{tntm_tbl:datasets}
    \begin{tabular}
    {
    l
    S[table-format=6.0]
    S[table-format=6.0]
    S[table-format=5.0]
    S[table-format=1.2]
    S[table-format=1.2]
    }
    \toprule
    Dataset & {Tweets} & {Authors} & 
    {Vocabulary} & {Words/Tweet} & {Hashtags/Tweet} \\
    \midrule
    T6          & 240517 &    150 &  5343 & 6.35 & 1.34 \\
    Generic     & 359478 & 213488 & 14581 & 6.84 & 0.10 \\
    Specific    & 214580 & 116685 & 15751 & 6.31 & 0.25 \\
    Events      & 107128 &  67388 & 12765 & 5.84 & 0.17 \\
    \bottomrule
    \end{tabular}
\end{table}

\subsection{Experiments and Results}
\label{tntm_sec:experiment}

We consider several tasks to evaluate the TNTM. The first task involves comparing
the TNTM with existing baselines on performing 
topic modelling on tweets.
We also compare the TNTM with the random function network model
on modelling the followers network.
Next, we evaluate the TNTM with ablation studies, 
in which we perform comparison with the TNTM itself but with each 
component taken away. 
Additionally, we evaluate the clustering performance of the TNTM, 
we compare the TNTM against the state-of-the-art tweets-pooling 
LDA method in \cite{Mehrotra:2013:ILT:2484028.2484166}.

\subsubsection{Experiment Settings}

In all the following experiments, 
we vary the discount parameters $\alpha$ for the topic distributions 
$\mu_0$\,, $\mu_1$\,, $\nu_i$\,, $\eta_m$\,, $\theta'_m$\,, and $\theta_m$\,,
we set $\alpha$ to $0.7$ for the word distributions $\psi$, $\phi'$ and $\gamma$ to induce 
power-law behaviour \citep{Goldwater:2011:PPD:1953048.2021075}. 
We initialise the concentration parameters $\beta$ to $0.5$, 
noting that they are learned automatically during inference, we set their 
hyperprior to $\mathrm{Gamma}(0.1, 0.1)$ for a vague prior.
We fix the hyperparameters $\lambda$, $s$, $l$ and $\sigma$ to $1$, 
as we find that their values have 
no significant impact on the model~performance.%
\footnote{
    We vary these hyperparameters over the range of $0.01$ to $10$ during testing.
}

In the following evaluations, we run the full inference
algorithm for 2,000 iterations for the models to converge.
We note that the MH algorithm only starts after 1,000 iterations.
We repeat 
each experiment five times to reduce the estimation error for the evaluations.

\subsubsection{Goodness-of-fit Test}

\begin{table}[t!]
    \centering
    \caption[
        Test Perplexity and Network Log Likelihood Comparisons
    ]{
        Test perplexity and network log likelihood comparisons 
        between the HDP-LDA, the nonparametric ATM,
        the random function network model and the TNTM.
        Lower perplexity indicates better model fitting.
        The TNTM significantly outperforms the other models
        in term of model fitting.
    }
    \label{tntm_tbl:perplexity}
    \begin{tabular}{rr@{\,\scriptsize$\pm$\,}lr@{\,\scriptsize$\pm$\,}l}
    \toprule
    Model & \multicolumn{2}{c}{Test Perplexity} & 
        \multicolumn{1}{r}{Network Log} & \!\!Likelihood 
    \\
    \midrule
    HDP-LDA & $ 840.03 $ & {\scriptsize $ 15.7 $} & \NAcell 
    \\ 
    Nonparametric ATM & $ 664.25 $ & {\scriptsize $ 17.76 $} & \NAcell 
    \\ 
    Random Function & \NAcell & $ -557.86 $ & {\scriptsize $ 11.2 $} 
    \\[2pt]
    \hdashline
    \noalign{\vskip 2.5pt} 
    TNTM & $ {\bf 505.01} $ & {\scriptsize $ 7.8 $} 
        & $ {\bf -500.63}$ & {\scriptsize $ 13.6 $} \\
    \bottomrule
    \end{tabular}
\end{table}

We compare the TNTM with the HDP-LDA and a nonparametric author-topic model (ATM) 
on fitting the text data (words and hashtags). 
Their performances are measured using perplexity on the test set 
(see Section~\ref{subsec:goodness-of-fit}).
The perplexity for the TNTM, accounting for both words and hashtags, is
\begin{align}
\mathrm{Perplexity}(\mathbf{W}^\circ) = 
\exp \! \Bigg(
-\frac{
    \log p \big(\mathbf{W}^\circ \,|\, \nu, \mu_1, \psi, \psi' \big)
}{
    \sum_{d=1}^D N_d + M_d
}
\Bigg) \,,
\end{align}

\noindent
where the likelihood $p \big(\mathbf{W}^\circ \,|\, \nu, \mu_1, \psi, \psi' \big)$ 
is broken into
\begin{align}
p \big(\mathbf{W}^\circ \,|\, \nu, \mu_1, \psi, \psi' \big)
= & \
\prod_{d=1}^D 
\prod_{m=1}^{M_d}
    p( y_{dm} \,|\, \nu, \mu_1, \psi' )    
\prod_{n=1}^{N_d}
    p( w_{dn} \,|\, y_d, \nu, \mu_1, \psi )    
\,.
\label{tntm_eq:word_conditional}
\end{align}

%

We also 
compare the TNTM against the original random function network model
in terms of the log likelihood of the network data, given by
$\log p(\mathbf{X} \,|\, \nu)$.
We present the comparison of the perplexity and the network log likelihood
in Table~\ref{tntm_tbl:perplexity}.
We note that for the network log likelihood, the less negative the better.
From the result, 
we can see that the TNTM achieves a much lower perplexity compared 
to the HDP-LDA and the nonparametric ATM.
Also, the nonparametric ATM is significantly better than the HDP-LDA.
This clearly shows that using more auxiliary information gives 
a better model fitting.
Additionally, 
we can also see that jointly modelling the text and network data leads 
to a better modelling on the followers network.

\subsubsection{Ablation Test}

\begin{table}[t!]
    \centering
    \caption[
        Ablation Test on the TNTM
    ]{
        Ablation test on the TNTM. The test perplexity and the 
        network log likelihood is evaluated on the TNTM against
        several ablated variants of the TNTM.
        The result shows that each component in the TNTM is important.
    }
    \label{tntm_tbl:ablation}
    \begin{tabular}{rr@{\,\scriptsize$\pm$\,}lr@{\,\scriptsize$\pm$\,}l}
    \toprule
    TNTM Model & \multicolumn{2}{c}{Test Perplexity} & 
        \multicolumn{1}{r}{Network Log} & \!\!Likelihood 
    \\
    \midrule
    No author & $ 669.12 $ & {\scriptsize $ 9.3 $} & \NAcell
    \\ 
    No hashtag & $ 1017.23 $ & {\scriptsize $ 27.5 $} & $ -522.83 $ & {\scriptsize $ 17.7 $} 
    \\ 
    No $\mu_1$ node & $ 607.70 $ & {\scriptsize $ 10.7 $} & $ -508.59 $ & {\scriptsize $ 9.8 $} 
    \\
    No $\theta'$-\,$\theta$ connection & $ 551.78 $ & {\scriptsize $ 16.0 $} & $ -509.21 $ & {\scriptsize $ 18.7 $}
    \\
    No power-law & $ 508.64 $ & {\scriptsize $ 7.1 $} & $ -560.28 $  & {\scriptsize $ 30.7 $} 
    \\[2pt]
    \hdashline
    \noalign{\vskip 2.5pt} 
    Full model & $ {\bf 505.01} $ & {\scriptsize $ 7.8 $} & $ {\bf -500.63}$ & {\scriptsize $ 13.6 $} \\
    \bottomrule
    \end{tabular}
\end{table}

Next, we perform an extensive ablation study with the TNTM. The components that are 
tested in this study are 
(1) authorship,
(2) hashtags,
(3) PYP $\mu_1$\,,
(4) connection between PYP $\theta'_d$ and $\theta_d$\,, and
(5) power-law behaviour on the PYPs.
We compare the full TNTM against variations in which each component is ablated.
%
%
%
Table~\ref{tntm_tbl:ablation} presents the test set perplexity and the 
network log likelihood of these models, it shows significant improvements 
of the TNTM over the ablated models.
From this, we see that
the greatest improvement on perplexity is from modelling the
hashtags, which suggests that the 
hashtag information is the most important for modelling tweets.
Second to the hashtags, the authorship information is very
important as well.
Even though modelling the power-law behaviour
is not that important for perplexity, we see that the
improvement on the network log likelihood is best achieved 
by modelling the power-law.
This is because the flexibility enables us to learn the 
author--topic distributions better, and thus allowing the TNTM to 
fit the network data better.
This also suggests that the authors in the corpus tend 
to focus on a specific topic rather than having a wide interest.

\subsubsection{Document Clustering and Topic Coherence}
\label{tntm_sec:clustering}

\begin{table}[t!]
    \centering
    \caption[
        Clustering Evaluations
    ]{
        Clustering evaluations of the TNTM against the LDA with different 
        pooling schemes. Note that higher purity and NMI indicate better 
        performance. The results for the different pooling methods 
        are obtained from Table 4 in \cite{Mehrotra:2013:ILT:2484028.2484166}.
        The TNTM achieves better performance on the purity and the NMI for 
        all datasets except for the Specific dataset, where it obtains the same
        purity score as the best pooling method.
    }
    \label{tntm_tbl:purity_nmi}
    \begin{tabular}{rcccccc}
    \toprule
    Method/Model & \multicolumn {3}{c}{Purity} & \multicolumn {3}{c}{NMI} 
    \\
    \midrule
    \textit{Data}
    & {\textit{Generic}} & {\textit{Specific}} & {\textit{Events}} 
    & {\textit{Generic}} & {\textit{Specific}} & {\textit{Events}} 
    \\
    \noalign{\vskip 2pt}
    No pooling & $ 0.49 $ & $ 0.64 $ & $ 0.69 $ & $ 0.28 $ & $ 0.22 $ & $ 0.39 $ 
    \\
    Author & $ 0.54 $ & $ 0.62 $ & $ 0.60 $ & $ 0.24 $ & $ 0.17 $ & $ 0.41 $ 
    \\
    Hourly & $ 0.45 $ & $ 0.61 $ & $ 0.61 $ & $ 0.07 $ & $ 0.09 $ & $ 0.32 $ 
    \\
    Burstwise & $ 0.42 $ & $ 0.60 $ & $ 0.64 $ & $ 0.18 $ & $ 0.16 $ & $ 0.33 $ 
    \\
    Hashtag & $ 0.54 $ & $ \mathbf{0.68} $ & $ 0.71 $ & $ 0.28 $ & $ 0.23 $ & $ 0.42 $
    \\
    \noalign{\vskip 2pt} 
    \hdashline
    \noalign{\vskip 2.5pt} 
    TNTM & $ \mathbf{0.66} $ & $ \mathbf{0.68} $ & $ \mathbf{0.79} $ 
    & $ \mathbf{0.43} $ & $ \mathbf{0.31} $ & $ \mathbf{0.52} $ 
    \\
    \bottomrule
    \end{tabular}
\end{table}

\cite{Mehrotra:2013:ILT:2484028.2484166} shows that running LDA on pooled tweets
rather than unpooled tweets gives significant improvement
on clustering. 
In particular, they find that grouping tweets based on
the hashtags provides most improvement.
Here, we show that instead of resorting to
such an \textit{ad hoc} method, the TNTM can achieve a 
significantly better result on clustering.
The clustering evaluations are measured with purity and normalised mutual information 
\citep[NMI, see][]{Manning:2008:IIR:1394399} described in \ref{subsubsec:clustering}.
Since ground truth labels are unknown,
we use the respective query terms as the ground truth
for evaluations. 
Note that tweets that satisfy multiple labels are removed.
Given the learned model, we assign a tweet to a cluster
based on its dominant topic.

We perform the evaluations on the Generic, Specific and Events datasets
for comparison purpose. 
We note the lack of network information in these datasets, and thus 
we employ only the HPYP part of the TNTM.
Additionally, since the
purity can trivially be improved by increasing the number of clusters, 
we limit the maximum number 
of topics to twenty for a fair comparison.
We present the results in Table~\ref{tntm_tbl:purity_nmi}.
We can see that the TNTM outperforms the pooling method in all 
aspects except on the Specific dataset, where it achieves the same purity 
as the best pooling scheme.

%

\subsubsection{Automatic Topic Labelling}
\label{tntm_sec:topicexplore}

Traditionally, researchers assign a topic 
for each topic--word distribution manually by inspection.
More recently, there have been attempts to label topics automatically in topic modelling.
For instance, \cite{Lau:2011:ALT:2002472.2002658} use Wikipedia to 
extract labels for topics, and \cite{Mehdad:2013:NAACL-HLT} use the entailment 
relations to select relevant phrases for topics. 
Here, we show that we can use hashtags 
to obtain good topic labels. 
In Table~\ref{tntm_tbl:exploration}, we display the top words 
from the topic--word distribution $\psi_k$ for each topic $k$.
Instead of manually assigning the topic labels,
we display the top three hashtags from the topic--hashtag distribution 
$\psi'_k$\,.
As we can see from Table~\ref{tntm_tbl:exploration}, 
the hashtags appear suitable as topic labels.
In fact, by empirically evaluating the 
suitability of the hashtags in representing the topics, we consistently find that,  
over 90\,\% of the hashtags are good candidates for the topic labels. 
Moreover, inspecting the topics show that the major hashtags coincide
with the query terms used in constructing the T6 dataset,
which is to be expected. 
This verifies that the TNTM is working properly.

\begin{table}[t!]
    \centering
    \caption[
        Topical Analysis
    ]{
        Topical analysis on the T6 dataset with the TNTM, 
        which displays the top three hashtags and the top $n$ 
        words on six topics.
        Instead of manually assigning a topic label to the
        topics, we find that the top hashtags can serve
        as the topic labels.
    }
    \label{tntm_tbl:exploration}
    \begin{tabular}{ccc}
        \toprule
         Topic & Top Hashtags & Top Words 
         \\
          \midrule
          \multirow{2}{*}{Topic 1} 
          & \multirow{2}{*}{{finance, money, economy}} 
         & finance, money, bank, marketwatch, \\
         && stocks, china, group, shares, sales
         \\
        \noalign{\vskip 2pt}
        \hdashline
        \noalign{\vskip 2.5pt} 
          \multirow{2}{*}{Topic 2} 
          & \multirow{2}{*}{{politics,\,iranelection,\,tcot}} 
         & politics, iran, iranelection, tcot, \\
         && tlot, topprog, obama, musiceanewsfeed
         \\
        \noalign{\vskip 2pt}
        \hdashline
        \noalign{\vskip 2.5pt} 
          \multirow{2}{*}{Topic 3} & \multirow{2}{*}{{music, folk, pop}} 
          & music, folk, monster, head, pop, \\
          && free, indie, album, gratuit, dernier 
          \\
        \noalign{\vskip 2pt}
        \hdashline
        \noalign{\vskip 2.5pt} 
          \multirow{2}{*}{Topic 4} & \multirow{2}{*}{{sports, women, asheville}} 
          & sports, women, football, win, game, \\
          && top, world, asheville, vols, team 
          \\          
        \noalign{\vskip 2pt}
        \hdashline
        \noalign{\vskip 2.5pt} 
          \multirow{2}{*}{Topic 5} & \multirow{2}{*}{{tech, news, jobs}} 
          & tech, news, jquery, jobs, hiring, \\
          && gizmos, google, reuters
          \\          
        \noalign{\vskip 2pt}
        \hdashline
        \noalign{\vskip 2.5pt} 
          \multirow{2}{*}{Topic 6} & \multirow{2}{*}{{science, news, biology}} 
          & science, news, source, study, scientists, \\
          && cancer, researchers, brain, biology, health
          \\
          \bottomrule
    \end{tabular}
\end{table}

%


\section{Conclusion}
\label{sec:conclusion}

In this article, we proposed a topic modelling framework utilising
PYPs, for which their realisation is a probability
distribution or another stochastic process of the same type.
In particular, for the purpose of performing inference, 
we described the CRP representation for the PYPs.
This allows us to propose a single framework, discussed in Section~\ref{sec:framework}, 
to implement these topic models,
where we modularise the PYPs (and other variables) into blocks that 
can be combined to form different models.
Doing so enables significant time to be saved on implementation of the
topic models.

We presented a general HPYP topic model, that can be seen as a
generalisation to the HDP-LDA \citep{TehJor2010a}.
The HPYP topic model is represented using a Chinese Restaurant Process (CRP)
metaphor 
\citep{TehJor2010a, Blei:2010:NCR:1667053.1667056, Chen:2011:STC:2034063.2034095},
and we discussed how the posterior likelihood of the HPYP topic model
can be modularised.
We then detailed the learning algorithm for the topic model 
in the modularised form.

We applied our HPYP topic model framework on Twitter data and 
proposed the Twitter-Network Topic model (TNTM).
The TNTM models the authors, text, hashtags, and the authors-follower
network in an integrated manner.
In addition to HPYP, the TNTM employs the Gaussian process (GP)
for the network modelling.
The main suggested use of the TNTM is for content discovery on social networks.
Through experiments, we show that jointly modelling of the text
content and the network leads to better model fitting 
as compared to modelling them separately.
Results on the qualitative analysis show that the learned
topics and the authors' topics are sound.
Our experiments suggest that
incorporating more auxiliary information leads to 
better fitting models.


\subsection{Future Research}
\label{sec:future}

For future work on TNTM, it would be interesting to apply TNTM
to other types of data, such as blogs and news feeds.
We could also use TNTM for other applications.
such as hashtag recommendation
and content suggestion for new Twitter users.
Moreover, we could extend TNTM
to incorporate more auxiliary information:
for instance, we can model the location of tweets and
the embedded multimedia contents such as URL, images and videos.
Another interesting source of information would be the path of 
retweeted content.%


Another interesting area of research is the combination of different
kinds of topic models for a better analysis. 
This allows us to transfer
learned knowledge from one topic model to another.
The work on combining LDA has already been looked at by 
\citet{Schnober:2015:CTM:2809936.2809939}, however, combining
other kinds of topic models, such as nonparametric ones,
is unexplored.


\acks{The authors like to thank Shamin Kinathil, the editors, 
and the anonymous reviewers for their valuable feedback
and comments.
NICTA is funded by the Australian Government through the Department of
Communications and the Australian Research Council through the ICT
Centre of Excellence Program.}


\appendix

\section{Posterior Inference for TNTM}

\subsection{Decrementing the Counts Associated with a Word or Hashtag}
\label{tntm_subsubsec:decrement}

When we remove a word or a hashtag during inference, we decrement by one the
customer count from the PYP associated with the word or the hashtag, 
that is, $c_k^{\theta_d}$ for word $w_{dn}$ ($z_{dn} = k$) and
$c_k^{\theta'_d}$ for hashtag $y_{dm}$ ($z'_{dm} = k$).
Decrementing the customer count may or may not decrement the respective table 
count. However, if the table count is decremented, then we would decrement
the customer count of the parent PYP.
This is relatively straight forward in Section~\ref{subsec:decrement}
since the PYPs have only one parent.
Here, when a PYP $\mathcal{N}$ has multiple parents, we would sample for one of 
its parent PYPs and decrement the table count corresponding to the parent PYP.
Although not the same, the rationale of this procedure follows Section~\ref{subsec:decrement}.

We explain in more details below.
When the customer count $c_k^\mathcal{N}$ is decremented,
we introduce an auxiliary variable $u_k^\mathcal{N}$ that indicates which
parent of $\mathcal{N}$ to remove a table from, or none at all.
The sample space for $u_k^\mathcal{N}$ is the $P$ parent nodes 
$\mathcal{P}_1\,, \dots, \mathcal{P}_P$ of $\mathcal{N}$, plus $\emptyset$.
When $u_k^\mathcal{N}$ is equal to $\mathcal{P}_i$\,,
we decrement the table count $t_k^{\mathcal{N} \to \mathcal{P}_i}$ 
and subsequently decrement the customer count $c_k^{\mathcal{P}_i}$
in node $\mathcal{P}_i $\,.
If $u_k^\mathcal{N}$ equals to $\emptyset$, we do not decrement any table count.
The process is repeated recursively as long as a customer count 
is decremented, that is, we stop when $u^\mathcal{N}_k = \emptyset$.

The value of $u^\mathcal{N}_k$ is sampled as follows:
\begin{equation}
p \big( u^\mathcal{N}_k \big) =
\left\{
    \begin{array}{ll}
        t^{\mathcal{N} \to \mathcal{P}_i}_k / c^\mathcal{N}_k 
            & ~~\mathrm{if ~~} u^\mathcal{N}_k = \mathcal{P}_i \\[6pt]
        1 - \sum_{\mathcal{P}_i} p \big( u^\mathcal{N}_k = \mathcal{P}_i \big)  
            & ~~\mathrm{if ~~} u^\mathcal{N}_k = \emptyset
        ~~.
    \end{array}
\right.
\label{tntm_eq:sampling_u}
\end{equation}

\noindent
To illustrate, when a word $w_{dn}$ (with topic $z_{dn}$) is removed, 
we decrement $c^{\theta_d}_{z_{dn}}$\,, 
that is, $c^{\theta_d}_{z_{dn}}$ becomes $c^{\theta_d}_{z_{dn}} - 1$. 
We then determine if this word contributes to any table in node $\theta_d$
by sampling $u^{\theta_d}_{z_{dn}}$ from Equation~\eqref{tntm_eq:sampling_u}.
If $u^{\theta_d}_{z_{dn}} = \emptyset$, we do not decrement any table count and 
proceed with the next step in Gibbs sampling; otherwise, 
$u^{\theta_d}_{z_{dn}}$ can either be $\theta'_d$ or $\eta_d$\,, 
in these cases, we would decrement $t^{\theta_d \to u^{\theta_d}_{z_{dn}}}_{z_{dn}}$ 
and $c^{u^{\theta_d}_{z_{dn}}}_{z_{dn}}$, and continue the process recursively.

We present the decrementing process in Algorithm~\ref{tntm_alg:decrement}.
To remove a word $w_{dn}$ during inference, we would need to decrement
the counts contributed by $w_{dn}$ (and $z_{dn}$).
For the topic side, we decrement the counts 
associated with node $\mathcal{N} = \theta_d$ with group $k = z_{dn}$ using 
Algorithm~\ref{tntm_alg:decrement}. 
While for the vocabulary side, we decrement the counts associated with
the node $\mathcal{N} = \psi_{z_{dn}}$ with group $k = w_{dn} $\,.
The effect of the word on the other PYP variables are implicitly considered
through recursion.

Note that the procedure to decrementing a hashtag $y_{dm}$ is similar,
in this case, we decrement the counts for $\mathcal{N} = \theta'_d$ with $k = z'_{dm}$ 
(topic side),
then decrement the counts for $\mathcal{N} = \psi'_{z'_{dm}}$ with $k = y_{dm}$
(vocabulary side).

\begin{algorithm}[t]
\vspace{2mm}
\caption[
    Decrementing Counts
]{
    Decrementing counts associated with a PYP $\mathcal{N}$ and group $k$.
}
\label{tntm_alg:decrement}
\begin{enumerate}
\item 
    Decrement the customer count $c_k^\mathcal{N}$ by one.
\item
    Sample an auxiliary variable $u_k^\mathcal{N}$ with Equation~\eqref{tntm_eq:sampling_u}.
\item
    For the sampled $u_k^\mathcal{N}$, perform the following:
    \begin{enumerate}
    \item 
        If $u_k^\mathcal{N} = \emptyset$, exit the algorithm.
    \item
        Otherwise, decrement the table count $t_k^{\mathcal{N} \to u_k^\mathcal{N}}$ by one
        and repeat Steps 2\,--\,4 by replacing $\mathcal{N}$ with $u_k^\mathcal{N} $.%
    \end{enumerate}
\end{enumerate}
\end{algorithm}
 
\subsection{Sampling a New Topic for a Word or a Hashtag}

After decrementing, we sample a new topic for the word or the hashtag. The sampling 
process follows the procedure discussed in Section~\ref{subsec:block_sampling}, 
but with different conditional posteriors (for both the word and the hashtag).
The full conditional posterior probability for the collapsed blocked Gibbs sampling 
can be derived easily. 
For instance, the conditional posterior for sampling the topic $z_{dn}$ of word $w_{dn}$ is
\begin{align}
 p(z_{dn}, \mathbf{T}, \mathbf{C} \, | \, 
  {\mathbf{Z}^{\circ}}^{-dn}, \mathbf{W}^\circ, 
  \mathbf{T}^{-dn}, \mathbf{C}^{-dn}, \mathbf{\Xi}) 
= 
\frac{
  p(\mathbf{Z}^\circ, \mathbf{T}, \mathbf{C} \, | \, \mathbf{W}^\circ, \mathbf{\Xi})
}{
  p({\mathbf{Z}^{\circ}}^{-dn}, \mathbf{T}^{-dn}, \mathbf{C}^{-dn} \, | \, 
    \mathbf{W}^\circ, \mathbf{\Xi})
} 
\label{tntm_eq:conditional_posterior}
\end{align}
which can then be easily decomposed into simpler form 
(see discussion in Section~\ref{subsec:block_sampling})
using Equation~\eqref{tntm_eq:hypy_posterior}. 
Here, the superscript $\Box^{-dn}$ indicates the word $w_{dn}$ and the topic $z_{dn}$ 
are removed from the respective sets. 
Similarly, the conditional posterior probability for sampling the topic $z'_{dm}$ 
of hashtag $y_{dm}$ can be derived as
\begin{align}
 p(z'_{dm}, \mathbf{T}, \mathbf{C} \, | \, 
  {\mathbf{Z}^{\circ}}^{-dm}, \mathbf{W}^\circ,
  \mathbf{T}^{-dm}, \mathbf{C}^{-dm}, \mathbf{\Xi}) 
= 
\frac{
  p(\mathbf{Z}^\circ, \mathbf{T}, \mathbf{C} \, | \, \mathbf{W}^\circ, \mathbf{\Xi})
}{
  p({\mathbf{Z}^{\circ}}^{-dm}, \mathbf{T}^{-dm}, \mathbf{C}^{-dm} \, | \,
    \mathbf{W}^\circ, \mathbf{\Xi})
} 
\label{tntm_eq:hashtag_conditional_posterior}
\end{align}
where the superscript $\Box^{-dm}$ signals the removal of the hashtag $y_{dm}$ and
the topic $z'_{dm}$\,.
As in Section~\ref{subsec:block_sampling}, we compute the posterior for all possible
changes to $\mathbf{T}$ and $\mathbf{C}$ corresponding to the new topic 
(for $z_{dn}$ or $z'_{dm}$). We then sample
the next state using a Gibbs sampler.

\subsection{Estimating the Probability Vectors of the PYPs with Multiple Parents}
\label{tntm_subsec:estimate_pyp}

Following Section~\ref{subsec:estimate_pyp}, we estimate the various probability
distributions of the PYPs by their posterior means.
For a PYP $\mathcal{N}$ with a single PYP parent $\mathcal{P}_1$\,, 
as discussed in Section~\ref{subsec:estimate_pyp}, we can estimate its 
probability vector $\hat{\mathcal{N}} = (\hat{\mathcal{N}}_1, \dots, \hat{\mathcal{N}}_K)$ as
\begin{align}
\hat{\mathcal{N}}_k
& =
\mathbb{E}[\mathcal{N}_k \,|\, \mathbf{Z}^\circ, \mathbf{W}^\circ, 
\mathbf{T}, \mathbf{C}, \mathbf{\Xi}]
\nonumber
\\[4pt]
& = 
\frac{
\big(\alpha^\mathcal{N} T^\mathcal{N} + \beta^\mathcal{N} \big) 
\mathbb{E}[\mathcal{P}_{1k} \,|\, \mathbf{Z}^\circ, \mathbf{W}^\circ, 
\mathbf{T}, \mathbf{C}, \mathbf{\Xi}]
+ c_k^\mathcal{N} - \alpha^\mathcal{N} T_k^\mathcal{N}
}{
\beta^\mathcal{N} + C^\mathcal{N}
}
~,
\label{tntm_eq:recover_vector}
\end{align}

\noindent
which lets one analyse the probability vectors in a topic model using recursion.

Unlike the above, the posterior mean is slightly more complicated for 
a PYP $\mathcal{N}$ that has multiple PYP parents 
$\mathcal{P}_1, \dots, \mathcal{P}_P$\,.
Formally, we define the PYP $\mathcal{N}$ as
\begin{align}
\mathcal{N} \,|\, \mathcal{P}_1, \dots, \mathcal{P}_P
\sim \mathrm{PYP}\Big(\alpha^\mathcal{N}, \beta^\mathcal{N}, 
    \rho_1^\mathcal{N} \mathcal{P}_1 + \dots + \rho_P^\mathcal{N} \mathcal{P}_P \Big) 
\,,
\end{align}

\noindent
where the mixing proportion 
$\rho^\mathcal{N} = (\rho_1^\mathcal{N}, \dots, \rho_P^\mathcal{N})$ 
follows a Dirichlet distribution with parameter 
$\lambda^\mathcal{N} = (\lambda_1^\mathcal{N}, \dots, \lambda_P^\mathcal{N})$:
\begin{align}
\rho^\mathcal{N} & \sim \mathrm{Dirichlet}\big(\lambda^\mathcal{N} \big) 
\,.
\end{align}

\noindent
Before we can estimate the probability vector, 
we first estimate the mixing proportion with its posterior mean
given the customer counts and table counts:
\begin{align}
\hat{\rho}_i^\mathcal{N}
=
\mathbb{E}[\rho_i^\mathcal{N} \,|\, \mathbf{Z}^\circ, \mathbf{W}^\circ, 
\mathbf{T}, \mathbf{C}, \mathbf{\Xi}]
=
\frac{
    T^{\mathcal{N} \to \mathcal{P}_i} + \lambda_i^\mathcal{N}
}{
    T^\mathcal{N} + \sum_i \lambda_i^\mathcal{N}
}
~.
\label{tntm_eq:estimate_rho}
\end{align}

\noindent
Then, we can estimate the probability vector 
$\hat{\mathcal{N}} = (\hat{\mathcal{N}}_1, \dots, \hat{\mathcal{N}}_K)$ by
\begin{align}
\hat{\mathcal{N}}_k
& = 
\frac{
\big(\alpha^\mathcal{N} T^\mathcal{N} + \beta^\mathcal{N} \big) 
\hat{H}_k^\mathcal{N}
+ c_k^\mathcal{N} - \alpha^\mathcal{N} T_k^\mathcal{N}
}{
\beta^\mathcal{N} + C^\mathcal{N}
}
~,
\label{tntm_eq:recover_vector_multi_parent}
\end{align}

\noindent
where $\hat{H}^\mathcal{N} = (\hat{H}_1^\mathcal{N}, \dots, \hat{H}_K^\mathcal{N})$ 
is the expected base distribution:
\begin{align}
\hat{H}_k^\mathcal{N} 
=
\sum_{i=1}^P 
    \hat{\rho}_i^\mathcal{N} 
    \mathbb{E}[\mathcal{P}_{ik} \,|\, \mathbf{Z}^\circ, \mathbf{W}^\circ, 
    \mathbf{T}, \mathbf{C}, \mathbf{\Xi}]
\,.
\end{align}


With these formulations, all the topic distributions and the word distributions 
in the TNTM
can be reconstructed from the customer counts and table counts. 
For instance, the author--topic distribution $\nu_i$ of each author $i$ can be determined 
recursively by first estimating the topic distribution $\mu_0$\,. 
The word distributions for each topic are similarly estimated.

\subsection{MH Algorithm for the Random Function Network Model}
\label{tntm_subsec:sampling_network}

Here, we discuss how we learn the topic distributions $\mu_0$ and $\nu$ from
the random function network model.
We configure the MH algorithm to start after running one thousand iterations
of the collapsed blocked Gibbs sampler, this is to we can quickly
initialise the TNTM with the HPYP topic model before running the full
algorithm.
In addition, this allows us to demonstrate the improvement to the TNTM
due to the random function network model.

To facilitate the MH algorithm, we have to represent the topic distributions
$\mu_0$ and $\nu$ explicitly as probability vectors, that is,
we do not store the customer counts and table counts for $\mu_0$ and $\nu$
after starting the MH algorithm.
In the MH algorithm, we propose new samples for $\mu_0$ and $\nu$, and then accept 
or reject the samples. 
The details for the MH algorithm is as follow.

In each iteration of the MH algorithm, 
we use the Dirichlet distributions as proposal distributions for $\mu_0$ and $\nu$:
\begin{align}
q(\mu_0^\mathrm{new} \,|\, \mu_0) & = \mathrm{Dirichlet}(\beta^{\mu_0} \mu_0)
\,,
\label{tntm_eq:propose_mu0}
\\[4pt]
q(\nu_i^\mathrm{new} \,|\, \nu_i) & = \mathrm{Dirichlet}(\beta^{\nu_i} \nu_i)
\,.
\label{tntm_eq:propose_nu}
\end{align}

\noindent
These proposed $\mu_0$ and $\nu$ are sampled given the their previous values,
and we note that the first $\mu_0$ and $\nu$ are computed using the technique
discussed in \ref{tntm_subsec:estimate_pyp}.
These proposed samples are subsequently used to sample $\mathbf{Q}^\mathrm{new}$.
We first compute the quantities $\varsigma^\mathrm{new}$ and
$\kappa^\mathrm{new}$ using the proposed $\mu_0^\mathrm{new}$ and $\nu^\mathrm{new}$ 
with Equation~\eqref{tntm_eq:mean_function} and 
Equation~\eqref{tntm_eq:kernel}.
Then we sample $\mathbf{Q}^\mathrm{new}$ given $\varsigma^\mathrm{new}$ and 
$\kappa^\mathrm{new}$ using the elliptical slice sampler (see \citealp{murray2009elliptical}):
\begin{align}
\mathbf{Q}^\mathrm{new} \sim \mathrm{GP}(\varsigma^\mathrm{new}, \kappa^\mathrm{new})
\,.
\label{tntm_eq:Q_new}
\end{align}


\noindent
Finally, we compute the acceptance probability $A' = \min(A, 1)$, where
\begin{align}
A = & \ 
\frac{
    p(\mathbf{Q}^\mathrm{new} \,|\, \mathbf{X}, \nu^\mathrm{new}, \mathbf{\Xi})
}{
    p(\mathbf{Q}^\mathrm{old} \,|\, \mathbf{X}, \nu^\mathrm{old}, \mathbf{\Xi})
}
\frac{
    f^*(\mu_0^\mathrm{new} \,|\, \nu^\mathrm{new}, \mathbf{T}) \, 
    \prod_{i=1}^A f^*(\nu_i^\mathrm{new} \,|\, \mathbf{T}) 
}{
    f^*(\mu_0^\mathrm{old} \,|\, \nu^\mathrm{old}, \mathbf{T}) \, 
    \prod_{i=1}^A f^*(\nu_i^\mathrm{old} \,|\, \mathbf{T})
}
\nonumber
\\
& \ \ \ \
\times
\frac{
    q(\mu_0^\mathrm{old} \,|\, \mu_0^\mathrm{new}) \, 
    \prod_{i=1}^A q(\nu_i^\mathrm{old} \,|\, \nu_i^\mathrm{new}) 
}{
    q(\mu_0^\mathrm{new} \,|\, \mu_0^\mathrm{old}) \, 
    \prod_{i=1}^A q(\nu_i^\mathrm{new} \,|\, \nu_i^\mathrm{old}) 
}
~,
\label{tntm_eq:acceptance}
\end{align}

\noindent
and we define $f^*(\mu_0 \,|\, \nu, \mathbf{T})$ and 
$f^*(\nu \,|\, \mathbf{T})$ as
\begin{align}
f^*(\mu_0 \,|\, \nu, \mathbf{T}) 
& =
\prod_{k=1}^K (\mu_{0k})^{t_k^{\mu_1} + \sum_{i=1}^A \nu_i}
~,
\\
f^*(\nu_i \,|\, \mathbf{T})
& =
\prod_{k=1}^K (\nu_{ik})^{\sum_{d=1}^D t_k^{\eta_d} I(a_d = i) }
~.
\label{tntm_eq:f_star_nu}
\end{align}

\noindent
The $f^*(\cdot)$ corresponds to the topic model posterior of
the variables $\mu_0$ and $\nu$ after we represent them as probability vectors
explicitly.
Note that we treat the acceptance probability $A$ as $1$ when the 
expression in Equation~\eqref{tntm_eq:acceptance} evaluates to more than $1$.
We then accept the proposed samples with probability $A$, if the sample
are not accepted, we keep the respective old values.
This completes one iteration of the MH scheme.
We summarise the MH algorithm in Algorithm~\ref{tntm_alg:network_mh}.

\begin{algorithm}[t!]
  \vspace{2mm}
  \caption{Performing the MH algorithm for one iteration.}
  \label{tntm_alg:network_mh}
  \begin{enumerate}
    \item
        Propose a new $\mu_0^\mathrm{new}$ with Equation~\eqref{tntm_eq:propose_mu0}.
    \item
        For each author $i$, propose a new $\nu_i^\mathrm{new}$ with
        Equation~\eqref{tntm_eq:propose_nu}.
    \item
        Compute the mean function $\varsigma^\mathrm{new}$ and the covariance
        matrix $\kappa^\mathrm{new}$ with Equation~\eqref{tntm_eq:mean_function}
        and Equation~\eqref{tntm_eq:kernel}.
    \item
        Sample $\mathbf{Q}^\mathrm{new}$ from Equation~\eqref{tntm_eq:Q_new}
        using the elliptical slice sampler from
        \citet{murray2009elliptical}.
    \item
        Accept or reject the samples with acceptance probability from
        Equation~\eqref{tntm_eq:acceptance}.
  \end{enumerate}
\end{algorithm}


\subsection{Hyperparameter Sampling}
\label{tntm_subsec:hyperparameter_sampling}

We sample the hyperparameters $\beta$ using an 
auxiliary variable sampler while leaving $\alpha$ fixed.
We note that the auxiliary variable sampler for PYPs that have
multiple parents are identical to that of PYPs with single parent,
since the sampler only used the total customer counts $C^\mathcal{N}$ and the 
total table counts $T^\mathcal{N}$ for a PYP $\mathcal{N}$.
We refer the readers to Section~\ref{subsec:hyperparameter} for details.

We would like to point out that hyperparameter sampling is
performed for all PYPs in TNTM for the first one thousand iterations.
After that, as $\mu_0$ and $\nu$ are represented as probability vectors
explicitly, we only sample the hyperparameters for the other PYPs (except 
$\mu_0$ and $\nu$).
We note that sampling the concentration parameters allows the topic distributions of 
each author to vary, that is, some authors have few very specific topics and some other 
authors can have a wider range of topics.
For simplicity, we fix the kernel hyperparameters $s$, $l$ and $\sigma$ 
to $1$. Additionally, we also make the priors for the mixing proportions 
uninformative by setting the $\lambda$ to~$1$.
We summarise the full inference algorithm for the TNTM in 
Algorithm~\ref{tntm_alg:inference}.
 
\begin{algorithm}[t!]
\vspace{2mm}
\caption{Full inference algorithm for the TNTM.}
\label{tntm_alg:inference}
    \begin{enumerate}
    \item 
        Initialise the HPYP topic model by assigning random topic 
        to the latent topic $z_{dn}$ associated with each word $w_{dn}$\,,
        and to the latent topic $z'_{dm}$ associated with each
        hashtag $y_{dm}$\,.
        Then update all the relevant customer counts $\mathbf{C}$ 
        and table counts $\mathbf{T}$.
    \item 
        For each word $w_{dn}$ in each document $d$, perform the following:
        \begin{enumerate}[label=(\alph*)]
        \item 
            Decrement the counts associated with $w_{dn}$ 
            (see \ref{tntm_subsubsec:decrement}).
           \item 
               Blocked sample a new topic for $z_{dn}$ and corresponding
               customer counts $\mathbf{C}$ and table counts $\mathbf{T}$ 
               (with Equation~\eqref{tntm_eq:conditional_posterior}).
           \item
               Update (increment counts) the topic model based on the sample.
           \end{enumerate}
    \item 
        For each hashtag $y_{dm}$ in each document $d$, perform the following:
        \begin{enumerate}[label=(\alph*)]
        \item 
            Decrement the counts associated with $y_{dm}$ 
            (see \ref{tntm_subsubsec:decrement}).
           \item 
               Blocked sample a new topic for $z'_{dn}$ and corresponding
               customer counts $\mathbf{C}$ and table counts $\mathbf{T}$ 
               (with Equation~\eqref{tntm_eq:hashtag_conditional_posterior}).
           \item
               Update (increment counts) the topic model based on the sample.
           \end{enumerate}
    \item 
        Sample the hyperparameter $\beta^\mathcal{N}$ for each PYP $\mathcal{N}$
        (see \ref{tntm_subsec:hyperparameter_sampling}).
    \item 
        Repeat Steps 2\,--\,4 for 1,000 iterations.
    \item
        Alternatingly perform the MH algorithm (Algorithm~\ref{tntm_alg:network_mh})
        and the collapsed blocked Gibbs sampler conditioned on $\mu_0$ and $\nu$.
    \item 
        Sample the hyperparameter $\beta^\mathcal{N}$ for each PYP $\mathcal{N}$
        except for $\mu_0$ and $\nu$.
    \item 
        Repeat Steps 6\,--\,7 until the model converges or when a fix number of iterations is reached.
    \end{enumerate}
\end{algorithm}

\bibliography{bibs/references}

\end{document}